\documentclass[lettersize,journal]{IEEEtran}
\usepackage{graphicx}
\usepackage{subcaption}
\usepackage{multirow}
\usepackage{booktabs}
\usepackage[table]{xcolor}
\usepackage{color}
\usepackage{colortbl}
\usepackage{tabularx}
\usepackage{bm}
\usepackage{url}
\usepackage{stackengine}
\usepackage{float}
\usepackage{verbatim}
\usepackage{array}
\usepackage{cite}
\usepackage{algorithm}
\usepackage{algorithmic}
\usepackage[colorlinks,linkcolor=cyan]{hyperref}
\usepackage{amsmath,amssymb} 
\usepackage{amsfonts}
\usepackage{comment}
\usepackage{svg}
\usepackage{orcidlink}
\usepackage{textcomp}
\usepackage{stfloats}
\usepackage[dvipsnames]{xcolor}             
\begin{document}

\title{ChessMamba: Structure-Aware Interleaving of State Spaces for Change Detection in Remote Sensing Images}

\author{Lei~Ding\orcidlink{0000-0003-0653-8373}, Tong~Liu, Xuanguang~Liu, Xiangyun~Liu, Haitao~Guo and Jun~Lu

\thanks{L. Ding is with the Information Engineering University, Zhengzhou, China, and also with the Aerospace Information Research Institute, Chinese Academy of Sciences, Beijing, China (E-mail: dinglei14@outlook.com).}

\thanks{Tong~Liu, Haitao~Guo, Xuanguang~Liu, Xiangyun~Liu, Haitao~Guo and Jun~Lu are with the Information Engineering University, Zhengzhou, China.}


\thanks{This document is funded by the Natural Science Foundation of China under Grant 42201443. (Corresponding author: Lei Ding.)}}

\markboth{Manuscript under review}%
{Shell \MakeLowercase{\textit{et al.}}: Bare Demo of IEEEtran.cls for IEEE Journals}

\maketitle

\begin{abstract}
Change detection (CD) in multitemporal remote sensing imagery presents significant challenges for fine-grained recognition, owing to heterogeneity and spatiotemporal misalignment. However, existing methodologies based on vision transformers or state-space models typically disrupt local structural consistency during temporal serialization, obscuring discriminative cues under misalignment and hindering reliable change localization. To address this, we introduce ChessMamba, a structure-aware framework leveraging interleaved state-space modeling for robust CD with multi-temporal inputs. ChessMamba integrates a SpatialMamba encoder with a lightweight cross-source interaction module, featuring two key innovations: (i) Chessboard interleaving with snake scanning order, which serializes multi-temporal features into a unified sequence within a single forward pass, thereby shortening interaction paths and enabling direct comparison for accurate change localization; and (ii) Structure-aware fusion via multi-dilated convolutions, selectively capturing center-and-corner neighborhood contexts within each mono-temporal. Comprehensive evaluations on three CD tasks, including binary CD, semantic CD and multimodal building damage assessment, demonstrate that ChessMamba effectively fuses heterogeneous features and achieves substantial accuracy improvements over state-of-the-art methods. The relevant code will be available at: \href{https://github.com/DingLei14/ChessMamba}{github.com/DingLei14/ChessMamba}. 
\end{abstract}


\section{Introduction}
\label{sec:intro}

Multi-source Remote Sensing (RS) imagery enables precise Change Detection (CD) across complex scenarios such as urban monitoring \cite{feng2023change}, disaster assessment \cite{zheng2021building, yao2024hierarchical}, and land cover dynamics \cite{ding2022bi, liu2025gstm}. However, the task of CD fundamentally requires disentangling meaningful changes from pervasive registration errors, seasonal variations, and modality-specific artifacts \cite{ding2025survey}. These discrepancies impede the accurate localization and semantic interpretation of fine-grained changes, attributed to computational challenges arising from the tight spatial coupling inherent in this spatio-temporal analysis.

Prior fusion methodologies, while evolving from convolutional architectures \cite{feng2023change}, recurrent networks \cite{chen2019change, mou2018learning} to self-attention designs \cite{chen2021remote, bandara2022transformer}, exhibit key operational limitations. Existing methodologies often struggle to harmonize global contextual modeling with precise localization, particularly as inputs scale to high resolutions. Vision Transformers (ViTs) mitigate local constraints but suffer quadratic complexity and remain vulnerable against spatial misalignment. While State Space Models (SSMs) like Mamba offer linear-time inference and long-range modeling, they remain challenged by directional biases and fine-grained feature integration in visual domains \cite{liu2025gstm}.

Efforts adapting SSMs to CD primarily focus on external architectural adjustments: VMamba introduces cross-scan modules to handle 2D data, and hybrid approaches combine SSMs with local CNNs. However, such designs retain intrinsic limitations. Multi-directional scanning strategies not only increase computational load but also impose inductive biases misaligned with the symmetric temporal nature in CD. Critically, existing SSM-based CD methods serialize inputs via predefined scanning orders \cite{chen2024rsmamba, chen2024changemamba} that disrupt 2D neighborhood coherence, obscuring multi-source disparities essential for reliable change localization. Crucially, literature approaches neglect internal refinement of state-space formulations for spatiotemporal coherence \cite{zhang2024cdmamba}, a gap limiting robust spatio-temporal feature fusion. These gaps hinder their ability to robustly fuse and contrast complex bitemporal features under misalignment, which is crucial to in CD tasks.

Motivated to fundamentally enhance the fusion paradigm within SSM computations, we introduce ChessMamba, a unified framework that integrates structural awareness into selective state-space modeling for robust CD with multisource inputs. Our core insight posits that robust cross-temporal interactions require enforcing locality through geometrically structured comparisons, a principle materialized via Chessboard Interleaving. This approach integrates two innovations: First, Chessboard Interleaving with bidirectional snake scanning explicitly decouples multi-temporal features within neighborhoods while ensuring topological consistency in the 1D sequence, reducing sensitivity to misregistration. Second, structure-aware fusion employs mono-context depthwise aggregation selectively preserving local context in each phase to reinforce cross-source contrast during state propagation. The SSM now processes sequences inherently refined for change localization, merging computational efficiency with discriminative power.

Our internal reconfiguration of SSMs elevates cross-source fusion from a supplemental operation to a fundamental architectural construct for robust change recognition. Crucially, the core contributions are:

\begin{itemize}
\item A Chessboard Shuffle module that interleaves bitemporal features with snake scanning, explicitly preserving 2D topology while shortening cross-source interaction paths;

\item A Mono-Contextual Aggregated State-Space Modeling (MCA-SSM) mechanism via dilated depthwise convolution prioritizes mono-temporal local contexts before state propagation, sharpening cross-source contrast;

\item Achieving State-of-the-art (SOTA) results in homogeneous and heterogeneous CD scenarios, affirming the effectiveness and generalization of proposed approach.
\end{itemize}

Extensive evaluations across three core CD tasks including binary (BCD), building damage assessment (BDA), and semantic change detection (SCD) demonstrate state-of-the-art results on three benchmark datasets (Levir-CD, Bright, and SECOND). It also demonstrates robust generalization to heterogeneous CD scenarios. 
\section{Related Work}
\label{sec:related}

\subsection{Change Detection}

Deep learning (DL) has revolutionized CD in RS, enabling robust, large-scale analysis. Binary change detection (BCD), the most-extensively studied CD task, identifies pixel-level changes between co-registered multi-temporal images to produce a changed/unchanged map. Traditional algebra or transformation-based methods are limited by hand-crafted features. with the rise of DL, the standard paradigm evolves to late-fusion Siamese CNN approach \cite{yang2021asymmetric}, using shared-weight encoders to extract temporal features before decoding the change. Recent architectures seek richer models of complex spatiotemporal relationships and better mitigation of pseudo-changes from illumination, season, or sensor variations by adopting Vision Transformers (ViT) \cite{bandara2022transformer} or hybrid CNN-Transformer models \cite{ding2024joint}.

Building Damage Assessment (BDA) extends CD into a fine-grained, multi-class task, critical for disaster response. Unlike BCD, BDA frameworks typically perform a dual-task: first localizing buildings, then classifying their structural integrity into multiple levels \cite{zheng2021building}, such as 'Survived' ('No damage'), 'Moderate' ('Minor/Major damage'), and 'Destroyed'. Methodologies often employ Siamese U-Net architectures to process pre- and post-event imagery \cite{adriano2021learning}. Recent approaches such as DamageFormer \cite{chen2022dual} leverage ViTs to better model non-local dependencies in the multitemporal data. Recent approaches also integrate multimodal data (Optical and SAR) to overcome weather constraints and enable rapid response \cite{chen2025bright}.

Semantic Change Detection (SCD) provides the richest information by identifying what has changed \cite{yang2021asymmetric}. Its core task is to identify "from-to" land-cover transitions (e.g., 'forest'$\rightarrow$'building'). Methodologies evolved from traditional post-classification comparison (comparing two independent segmentation maps) to more robust, end-to-end joint frameworks \cite{zheng2022changemask}. These modern approaches, often using Siamese encoders with advanced backbones like ViTs or Mamba \cite{mei2024scd, liu2025gstm}, perform bi-temporal semantic reasoning \cite{ding2022bi} or joint spatio-temporal modeling \cite{ding2024joint} to directly predict the transition labels, offering greater accuracy by avoiding error propagation.

\subsection{Visual Mamba}

The paradigm shift in computer vision from CNNs/ViTs to Mamba architectures addresses critical limitations: CNNs have limited receptive fields, while ViTs suffer prohibitive quadratic complexity with high-resolution data. SSMs, notably Mamba \cite{gu2023mamba} with its Selective State Space (S6) mechanism, offer linear complexity and robust long-range modeling. S6's input-dependent parameterization enables selective information flow crucial for long visual sequences.

Early visual adaptations include Vision Mamba (ViM) \cite{zhu2024vision}, integrating bidirectional SSM and ViT-like positional embeddings, and VMamba \cite{liu2024vmamba}, introducing a Cross-Scan Module using four-way scanning and depth-wise convolutions for 2D image adaptation. Subsequent research refined 2D-to-1D conversion: PlainMamba \cite{yang2024plainmamba} pursued continuous 2D scanning; LocalMamba \cite{huang2024localmamba} employed windowed selective scanning to enhance local features. Spatial-Mamba \cite{xiao2024spatial} proposed a direct state-space solution via structure-aware state fusion with dilated convolutions. This method integrates local spatial dependencies within a unidirectional scan, achieving efficiency and structural awareness without multi-scan directional bias.

\begin{figure*}[t]
\centering
    \includegraphics[width=0.9\linewidth]{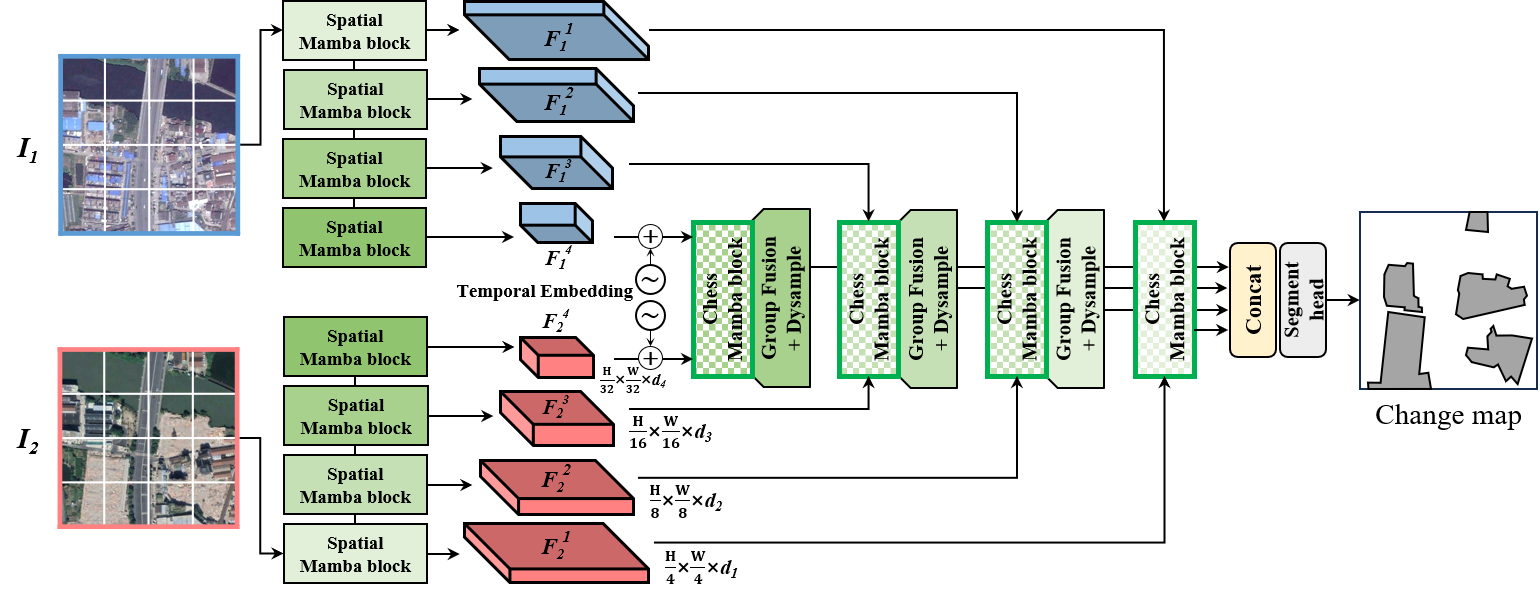}
    \caption{Overview of ChessMamba, a spatio-temporal context-aware SSM framework for feature fusion. The Chessboard Mamba blocks facilitate cross-temporal interactions within state-space propagation, sharpening change localization under misalignment.}
\label{fig.network}
\end{figure*}

\subsection{Mamba in Remote Sensing}

The ViM offers significant advantages for for RS tasks, as its linear computational complexity and long-range dependency modeling efficiently manages the challenges of high-resolution and multi-temporal data. This overcomes the inherent limitations of CNNs (restricted receptive fields) and Transformers (quadratic complexity) for large RS scenes \cite{bao2025vision}.

Initial foundational models rapidly evolved for specific tasks. For instance, RSMamba \cite{chen2024rsmamba} employs a dynamic multi-path gating mechanism for RS image classification. This capability quickly transitioned to dense prediction. In semantic segmentation (SS), Samba and RS$^3$Mamba \cite{zhu2024samba, ma2024rs3mamba} demonstrated high accuracy by utilizing pure Mamba backbones or hybrid architectures to effectively fuse global context captured by Mamba with local spatial features, a crucial step for accurate pixel-level tasks on massive RS images.

The CD task with multi-source inputs demands efficient bitemporal interactions. ChangeMamba \cite{chen2024changemamba} pioneered the application of a Visual Mamba encoder for CD tasks by incorporating specialized Mamba-based spatio-temporal modeling mechanisms within its decoder for efficient temporal feature interaction. Recognizing Mamba's inherent limitation in capturing fine-grained local detail, many subsequent works focused on robust hybridization: CDMamba \cite{zhang2024cdmamba} and ConMamba \cite{dong2024conmamba} proposed combining Mamba's global strength with convolution's ability to extract local clues using specialized hybrid blocks, while VMI-CD \cite{zhang2025enhancing} enhanced the VMamba backbone with lightweight feature interaction modules for efficiency. Addressing multi-temporal sequences, GSTM-SCD \cite{liu2025gstm} introduced a graph-enhanced spatio-temporal SSM to better capture dynamic changes in SCD. 

Overall, existing Mamba efforts in RS and CD primarily focus on external optimizations, e.g., specialized scanning schemes or hybrid CNN-Mamba architectures. Consequently, research dedicated to internally enhancing the formulation of state space for spatio-temporal modeling of multi-source features worth further exploration.

\section{Methodology}
\label{sec:method}

\subsection{Overall Architecture}

Classical SSMs evolve a latent state with linear dynamics and emit observations via
\begin{equation}
    x_t = \overline{A}_t x_{t-1} + \overline{B}_t u_t, \quad y_t = C_t x_t + D u_t.
\end{equation}
where $u_t$ and $y_t$ are the input and output states, $x_t$ is the hidden state at step $t$,  $\overline{A}_t, \overline{B}_t$ are the discretized parameters, $C_t$ and $D$ are two projection matrices. Mamba instantiates a gated and input-selective parameterization of the continuous-time SSM and discretizes it to enable linear-time inference on long sequences while preserving strong long-range dependency modeling \cite{liu2024vmamba}. Concretely, selective input and gating terms modulate the effective state transition and input matrices, yielding an efficient scan that scales as $O(N)$ with sequence length.

Spatial-Mamba extends this principle from 1D sequences to 2D feature maps by reinterpreting the spatial tensor as scan-friendly sequences and interleaving lightweight local operators (e.g., depthwise conv and normalization) for stability \cite{xiao2024spatial}. Given bi-temporal inputs $(I_1,I_2)$, our encoder first adopts SpatialMamba to extract a multi-scale hierarchy $\{F_{1}^s\}_{s=1}^{S}$.

Our motivation is to preserve the computational virtues of Spatial-Mamba while inserting a principled, structure-aware fusion stage that both contrasts and aligns heterogeneous evidence before the scan. Concretely, we use SpatialMamba as an encoder to extract multi-scale features from each phase (i.e., the distinct time point in RS acquisitions). The core novelty lies in the ChessMamba decoder: at each scale we perform lightweight cross-time enhancement, then align every scale to the highest resolution and concatenate them, finally applying a single-stage reduction-and-refinement. This organization keeps the scan linear and memory-friendly, while ensuring the decoder presents the SSM with sequences that are already contrastive and geometrically consistent.

The decoder is organized with stacked ChessMamba blocks and upscaling layers. First the ChessMamba block first performs cross-time enhancement at each scale, then aligns all scales to the highest resolution and concatenates them along channels, followed by a single-stage refinement to produce the final change representation. To suppress interpolation artifacts when upsampling irregular-sized feature maps, all upsampling uses a DySample operator \cite{liu2023learning} with a bilinear fallback for exact size alignment.

Conceptually, the architecture turns multi-source fusion from a side utility into the central design principle of the decoding stage. By shaping how evidence is combined before the SSM, we obtain long-range modeling that is not merely agnostic to source relationships, but actively benefits from them, delivering more decisive and stable change representations.

\begin{figure*}[t]
\centering
    \includegraphics[width=0.81\linewidth]{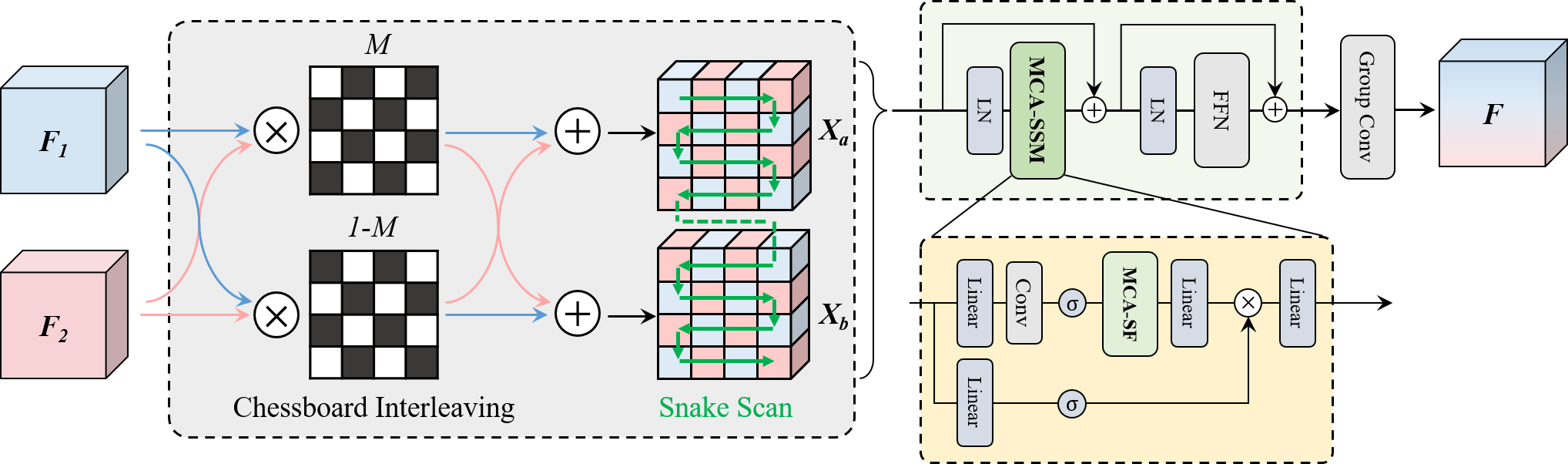}
    \caption{Calculations within a Chess-Mamba block. The chessboard interleaving enables direct per-pixel comparisons while preserving 2D neighborhood topology.}
\label{fig.chessmamba}
\end{figure*}

\subsection{Chessboard Shuffle}

Prior multi-source fusion strategies often rely on uniform stacking or pairwise attention, which can mix temporal neighborhoods and reduce discriminability in the presence of spatial shift, illumination drift, or modality-specific artifacts. The consequence is a scan path that does not respect how spatial neighborhoods differ across sources, making the 1D sequence a poor surrogate for the 2D topology that actually governs cross-source agreement and disagreement.

We are motivated to impose a geometric prior that preserves “in-phase” neighborhoods and magnifies cross-source contrast in a simple, model-agnostic way. Chessboard shuffle constructs two mirror composites with a binary chessboard mask and its complement, ensuring each spatial location is dominated by same-color neighbors while its counterpart captures the complementary phase. To explicitly align "same-phase" spatial neighborhoods and mitigate temporal aliasing before sequence modeling, we introduce chessboard shuffle. Let $M \in \{0, 1\}^{H \times W}$ be a binary chessboard mask (black–white alternation) and $I$ be the all-ones matrix. From a cross-time fused feature pair at a given scale $s$ (e.g., group-wise fusion applied to $\{F_{1}^s\}$ and $\{F_{2}^s\}$), we construct two mirror streams:

\begin{align}
    X^{a}=M\odot F_{1}^s+(I-M)\odot F_{2}^s,\\
    \quad X^{b}=(I-M)\odot F_{1}^s+M\odot F_{2}^s.
\end{align}

These two streams cover complementary chessboard cells such that each position is surrounded by 'cross-phase' neighbors. We then interleave them along the width dimension and obtain a scan-ready tensor of shape $(B,H,2W,C)$. A row-wise zigzag (snake) order is used to construct a 1D path: even rows are scanned left-to-right while odd rows are flipped horizontally, ensuring continuity between row ends and aligns the 1D trajectory with the 2D chessboard topology. This layout makes the subsequent 1D SSM scan topologically consistent with the 2D chessboard prior, preserving cross-time saliency while suppressing aliasing and favoring stable state propagation. 

In effect, chessboard shuffle reframes fusion as a structured reindexing problem: it injects an explicit, spatially alternating prior before sequence modeling. The SSM hence receives a topologically consistent path where in-phase context is preserved and cross-phase contrast is accentuated, improving both stability and discriminative power without incurring the overhead of heavy cross-attention.

\subsection{Mono-Context Aggregated SSM}

Local pre-aggregation before sequence modeling is common in vision SSMs, yet standard isotropic kernels or unconstrained mixing may inadvertently couple orthogonal (cross-phase) neighbors and blur the very cues that distinguish sources. This issue compounds in multi-source settings, where the scan’s state can carry forward mixed-phase noise rather than coherent, in-phase evidence.

Our motivation is to encode a mono-context bias that privileges same-color neighborhoods at the very entrance to the SSM. Before the SSM scan, we apply a Mono-Context Aggregated State Fusion (MCA-SF) mechanism to emphasize prioritize phase-aligned evidence before state propagation. Specifically, MCA-SF extends the structure-aware state fusion principle from SpatialMamba \cite{xiao2024spatial} through multi-dilated processing. For each embedded feature variable $x_t$ at the position $t$, we compute aggregated states:
\begin{equation}
    h_{t} = \sum \mathrm{DWConv}_d\left(x_t, \mathbf{K}_d\right), \text{s.t.} \mathbf{K}_1 = \begin{bmatrix} k_1 & 0 & k_3 \\ 0 & k_5 & 0 \\ k_7 & 0 & k_9 \end{bmatrix}
\end{equation}
where $\mathbf{K}_d \in \mathbb{R}^{C\times1\times3\times3}$ is the depthwise kernel, learnable weights $k_i$ remain only at center and corner positions for $d=1$, while $d\in\{3,5\}$ use full kernels. 

The aggregated state $h_{t}$ is then fed into the selective SSM block:
\begin{equation}
    x_t = \overline{A}_t x_{t-1} + \overline{B}_t u_t, \quad y_t = C_t h_t + D u_t.
\label{mca-sf}
\end{equation}

This MCA-SF mechanism combines multi-scale receptive fields to capture local structural dependencies, while restricting aggregation to mono-temporal neighbors. This preserves phase-aligned spatial integrity (e.g., edges, textures) during fusion, avoiding disruption to localized semantics. By unifying global state transitions with locally coherent spatial information within each mono-temporal, the fused state $h_t$ enriches representations and sharpens modeling of structural semantic transitions (i.e., changes).

The structure-aware state variable $h_t$ incorporates additional neighboring state variables. By considering both the global long-range and the mono-temporal local spatial information, the fused state variable $h_t$ gains a richer spatio-temporal context, leading to improved adaptability and a more comprehensive understanding of the semantic transitions (i.e., changes).

The resulting 1D feature sequences are reshaped into their original 2D configurations for both time phases. These aligned 2D representations are then directly summed and fused. This additive fusion leverages the inherent positional complementarity of the chessboard pattern, where spatially adjacent features from the two phases exhibit non-overlapping coverage, thereby enabling efficient integration of multi-temporal contexts without explicit alignment constraints. The fused features retain structural integrity for subsequent hierarchical processing.

The fused multi-scale features are then fused using grouped $1\times1$ convolution with a lightweight depthwise $3\times3$ smoothing, and all enhanced scales $\{\widetilde{F}^{(s)}\}_{S=1}$ are resized to the highest resolution and concatenated, followed by a refinement:
\begin{equation}
    F_{\mathrm{cat}} = \{ \uparrow\widetilde{F}^1, \ldots, \uparrow\widetilde{F}^S\}, F_{\mathrm{chg}} = \phi_{3\times 3}(\phi_{1\times 1}(F_{\mathrm{cat}})),
\end{equation}
where $\phi_{1\times 1}$ reduces channel dimensionality and $\phi_{3\times 3}$ performs spatial optimization, $\uparrow$ is the upsample operation implemented with Dysample. Combined with the interleaved, zigzag sequence, the SSM now propagates state over tokens that are already filtered for in-phase coherence and arranged to emphasize cross-source contrast.

This MCA-SF mechanism enhances cross-position interactions without compromising local discriminability, leading to enriched representations of structural details. The result is more stable long-range propagation, stronger localization of changes, and improved robustness to trivial variation.

\begin{figure}[t]
\centering
    \includegraphics[width=0.8\linewidth]{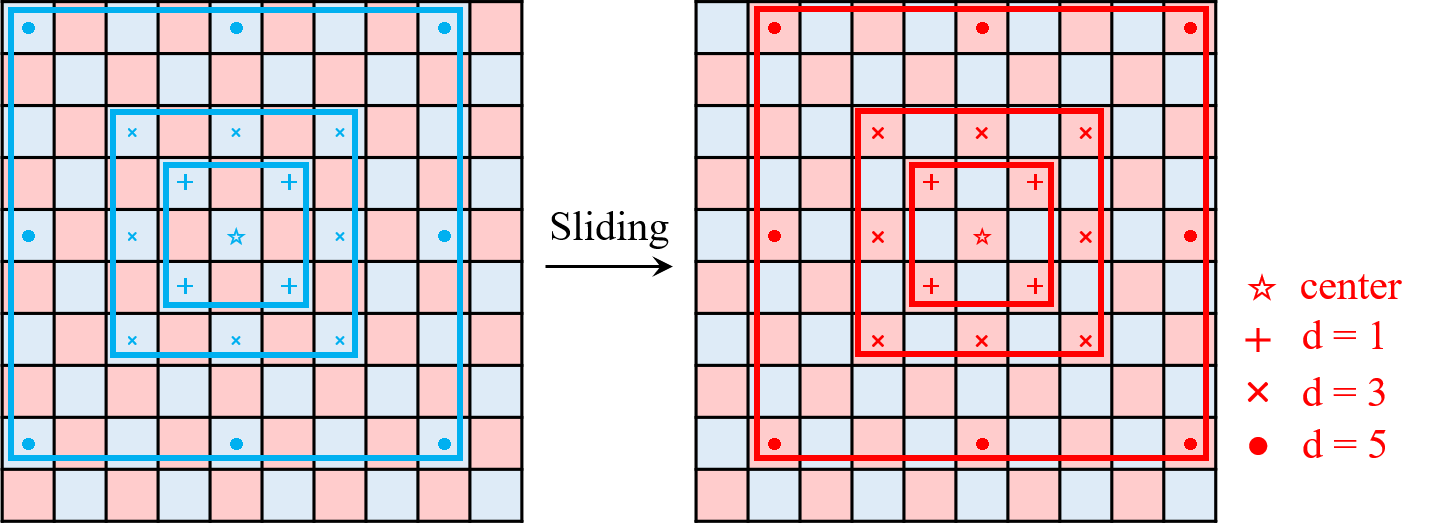}
    \caption{Aggregation of Mono-Context. At each position, dilated kernels aggregate local context exclusively from one source.}
\label{fig.neibor}
\end{figure}

\subsection{Task Adaptation and Training Details}
\label{sc3.tasks}

Prior CD works that adopt Mamba encoders typically treat multi-source inputs by late concatenation or generic attention mixing, leaving cross-source comparison and alignment under-specified. This often weakens supervision signals: binary change loses semantic context, semantic change struggles to couple with change cues, and multi-class damage suffers from heterogeneous sensing gaps (e.g., optical vs. SAR). Our motivation is to make the decoder the locus of fusion, such that the scan receives sequences that are already contrastive across sources and contextually aligned in space. Concretely, we instantiate two ChessMamba variants for BDA and SCD tasks. In experiments, BCD and SCD use homogeneous inputs (two optical images) and therefore share encoder weights; BDA uses heterogeneous inputs (optical + SAR), so the two encoders are not weight-sharing. This architectural choice preserves representational symmetry when appropriate and decouples sensing-specific biases when necessary.

\begin{figure}[t]
\centering
    \includegraphics[width=1\linewidth]{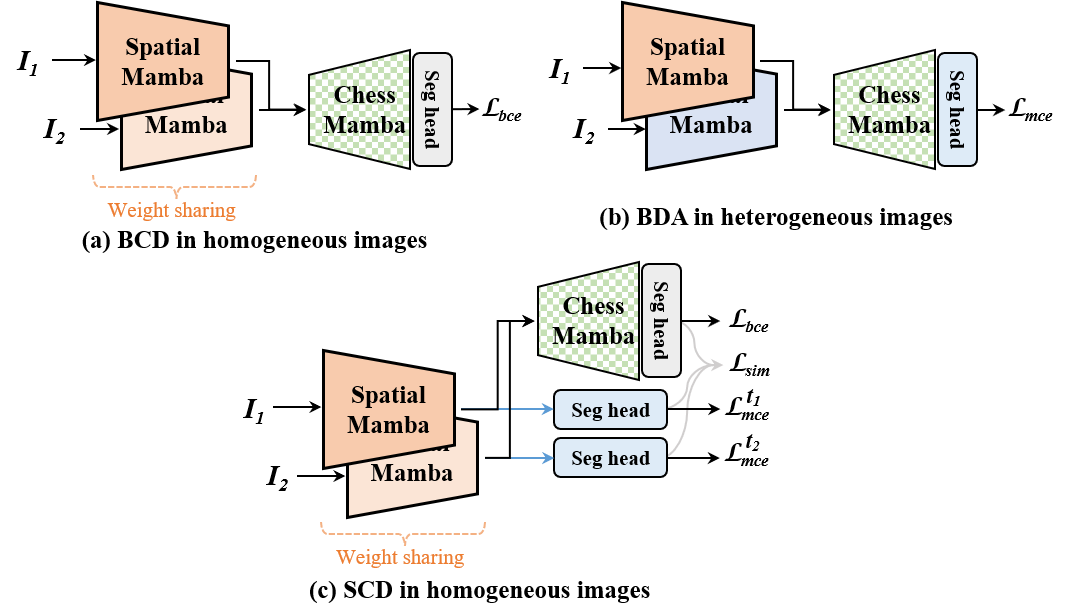}
    \caption{Variants of ChessMamba adapting to different CD tasks.}
\label{fig.neibor}
\end{figure}

For optimization, training objectives are aligned with each task's output structure. BCD is trained with a binary cross-entropy loss:
\begin{equation}
    \mathcal{L}_{\mathrm{bce}}=-\frac{1}{N} \sum_{i=1}^{N}\left[y_{i} \log p_{i}+\left(1-y_{i}\right) \log \left(1-p_{i}\right)\right]
\end{equation}
where $N$ denotes the total number of pixels, $p_i$ is the predicted change probability at pixel $i$.

BDA is a multi-class task and follows a semantic segmentation objective with multi-class cross-entropy:
\begin{equation}
    \mathcal{L}_{\mathrm{mce}}=-\frac{1}{N} \sum_{i=1}^{N} \sum_{c=1}^{C} \mathbf{1}\left[y_{i}=c\right] \log p_{i, c}.
\end{equation}
where $\mathbf{1}[\cdot]$ is a indicator function acts as a binary class selector.

SCD jointly predicts change and per-phase semantics, thus we combine binary and multi-class terms with an auxiliary semantic consistency term:
\begin{equation}
    \mathcal{L}_{\mathrm{scd}} = \mathcal{L}_{\mathrm{bce}} +  (\mathcal{L}_{\mathrm{mce}}^{t_1} + \mathcal{L}_{\mathrm{mce}}^{t_2})/2 + \mathcal{L}_{\mathrm{sim}},
\end{equation}
where $\mathcal{L}_{sim}$ promotes semantic agreement over unchanged regions, implemented following the literature practice \cite{ding2022bi, ding2024joint}.

We adopt AdamW with an initial learning rate of $5e^{-4}$, warm-up over the first 5 epochs, and exponential decay thereafter. The training is conducted for 50 epochs, utilizing a batch size of 6 with spatial augmentations including random cropping with $512\times512$ windows, and horizontal/vertical flipping. Input RS images are independently normalized using their channel-wise means. During validation and testing on high-resolution imagery (1024$\times$1024 pixels), we deploy sliding-window inference with non-overlapping tiles and test-time augmentation (flips). Final predictions are stitched into whole-image annotations for metric computation. This protocol preserves task-specific data distributions while enabling the decoder’s chessboard structure and mono-contextual priors to consistently enhance cross-temporal contrast and stabilize state propagation across BCD, SCD, and BDA tasks.

\section{Experiments}
\label{sec:method}

\subsection{Datasets and Evaluation Metrics}

We conduct experiments leveraging three pivotal RS benchmarks: Levir for BCD, BRIGHT for BDA, and SECOND for SCD. Levir pioneered large-scale, high-resolution urban change analysis, facilitating BCD model evolution. BRIGHT contributes multimodal (optical/SAR) disaster-responsive data, enabling all-weather BDA critical for time-sensitive applications. SECOND further advances SCD with fine-grained land cover transitions, supporting complex change analysis. Detailed specifications of these datasets are provided in Table \ref{Table.Datasets}.

\begin{table}[h]
    \centering
    \caption{Statistical summary of the experimental datasets.} \label{Table.Datasets}
    \resizebox{1\linewidth}{!}{%
    \begin{tabular}{c|c|c|c|r|l}
    \toprule
    \rowcolor[HTML]{EFEFEF} 
        Datasets & Sensors & GSD & Image size & Data Size & \multicolumn{1}{c}{Change Type} \\
        \hline
        Levir & optical & 0.5m & 1024×1024 & 637 pairs & building change \small (2 classes) \\
        BRIGHT & optical + SAR & 0.3-1m & 1024×1024 & 4,246 pairs & building damage \small (4 classes) \\
        SECOND & optical & 0.5-3m & 512×512 & 4,662 pairs & land cover \small (6 'from-to' classes)\\
    \bottomrule
    \end{tabular}}
\label{table.datasets}
\end{table}

Notably, ablation studies focus on Levir for BCD due to its role as the fundamental change detection task, isolating the assessment of change extraction precision. Comparative experiments benchmark our approach against SOTA methods across all three datasets.

We employ the most frequently adopted metrics in CD tasks for accuracy assessment, including Overall Accuracy (OA), Precision (Pre), Recall (Rec), Intersection over Union (IoU) and $F_1$ score (harmonic mean of precision/recall) \cite{bandara2022transformer, ding2024adapting}. In BDA, $F_1^{loc}$ evaluates building localization accuracy, while $F_1^{clf}$ measures damage classification correctness \cite{chen2025bright}. In SCD, Separated Kappa (SeK) addresses label imbalance by decoupling change categories \cite{yang2021asymmetric}, and $F_{scd}$ (F-score for semantic change) assesses semantic segmentation accuracy in change regions \cite{ding2024joint}.

\subsection{Ablation Studies}

We systematically evaluate component contributions on the Levir dataset in Table \ref{Table.Ablation}. The baseline (Spatial Mamba encoder + UPerNet head) achieves 83.26\% IoU. Introducing Chessboard Shuffle to the decoder enhances neighborhood alignment (+0.71\% IoU), while MCA-SSM aggregation alone boosts robustness (+1.04\% IoU). The unified ChessMamba decoder synergizes both components, achieving 85.20\% IoU (+1.94\%), a dominant improvement among ablated variants. Critically, ChessMamba drives the largest recall gain (+2.09\%), validating its efficacy in reducing missed detections of subtle changes. Notably, compared to Mamba-based alternatives (ChangeMamba \cite{chen2024changemamba} and RSMamba \cite{chen2024rsmamba}), our method demonstrates superior localization precision.

\begin{table}[h!]
\centering
\caption{Ablation study on the Levir-CD dataset.}
\resizebox{1\linewidth}{!}{%
\begin{tabular}{l|ccccc}
\toprule
\multirow{2}{*}{Methods} & \multicolumn{5}{c}{Accuracy (\%)}\\
\cmidrule(lr){2-6}
& $OA$ & $Pre$ & $Rec$ & $F_1$ & IoU \\
\midrule
ChangeMamba      & 99.04 & 90.73 & 90.30 & 90.51 & 82.67 \\
RSMamba          & 99.05 & 90.70 & 90.63 & 90.67 & 82.93 \\
\midrule
SpatialMamba-CD (\textit{baseline})  & 99.09 & 93.25 & 88.59 & 90.86 & 83.26 \\
+ Chessboard shuffle only  & 99.14 & 93.41 & 89.25 & 91.28 & 83.97 \\
\multicolumn{1}{r|}{\textit{$\Delta$ vs. baseline}}      &
\multicolumn{1}{r}{\textcolor{ForestGreen}{\small $\uparrow$ 0.05}} &
\multicolumn{1}{r}{\textcolor{ForestGreen}{\small $\uparrow$ 0.16}} &
\multicolumn{1}{r}{\textcolor{ForestGreen}{\small $\uparrow$ 0.66}} &
\multicolumn{1}{r}{\textcolor{ForestGreen}{\small $\uparrow$ 0.42}} &
\multicolumn{1}{r}{\textcolor{ForestGreen}{\small $\uparrow$ 0.71}} \\
+ MCA-SSM only  & 99.16 & 93.30 & 89.73 & 91.49 & 84.30 \\
\multicolumn{1}{r|}{\textit{$\Delta$ vs. baseline}}      &
\multicolumn{1}{r}{\textcolor{ForestGreen}{\small $\uparrow$ 0.07}} &
\multicolumn{1}{r}{\textcolor{ForestGreen}{\small $\uparrow$ 0.05}} &
\multicolumn{1}{r}{\textcolor{ForestGreen}{\small $\uparrow$ 1.14}} &
\multicolumn{1}{r}{\textcolor{ForestGreen}{\small $\uparrow$ 0.63}} &
\multicolumn{1}{r}{\textcolor{ForestGreen}{\small $\uparrow$ 1.04}} \\
+ ChessMamba & 99.20 & 93.37 & 90.68 & 92.01 & 85.20 \\
\multicolumn{1}{r|}{\textit{$\Delta$ vs. baseline}}     &
\multicolumn{1}{r}{\textcolor{ForestGreen}{\small $\uparrow$ 0.11}} &
\multicolumn{1}{r}{\textcolor{ForestGreen}{\small $\uparrow$ 0.12}} &
\multicolumn{1}{r}{\textcolor{ForestGreen}{\small $\uparrow$ 2.09}} &
\multicolumn{1}{r}{\textcolor{ForestGreen}{\small $\uparrow$ 1.15}} &
\multicolumn{1}{r}{\textcolor{ForestGreen}{\small $\uparrow$ 1.94}} \\
\bottomrule
\end{tabular}%
}
\label{Table.Ablation}
\end{table}

\begin{figure}[h]
\centering
    \includegraphics[width=1\linewidth]{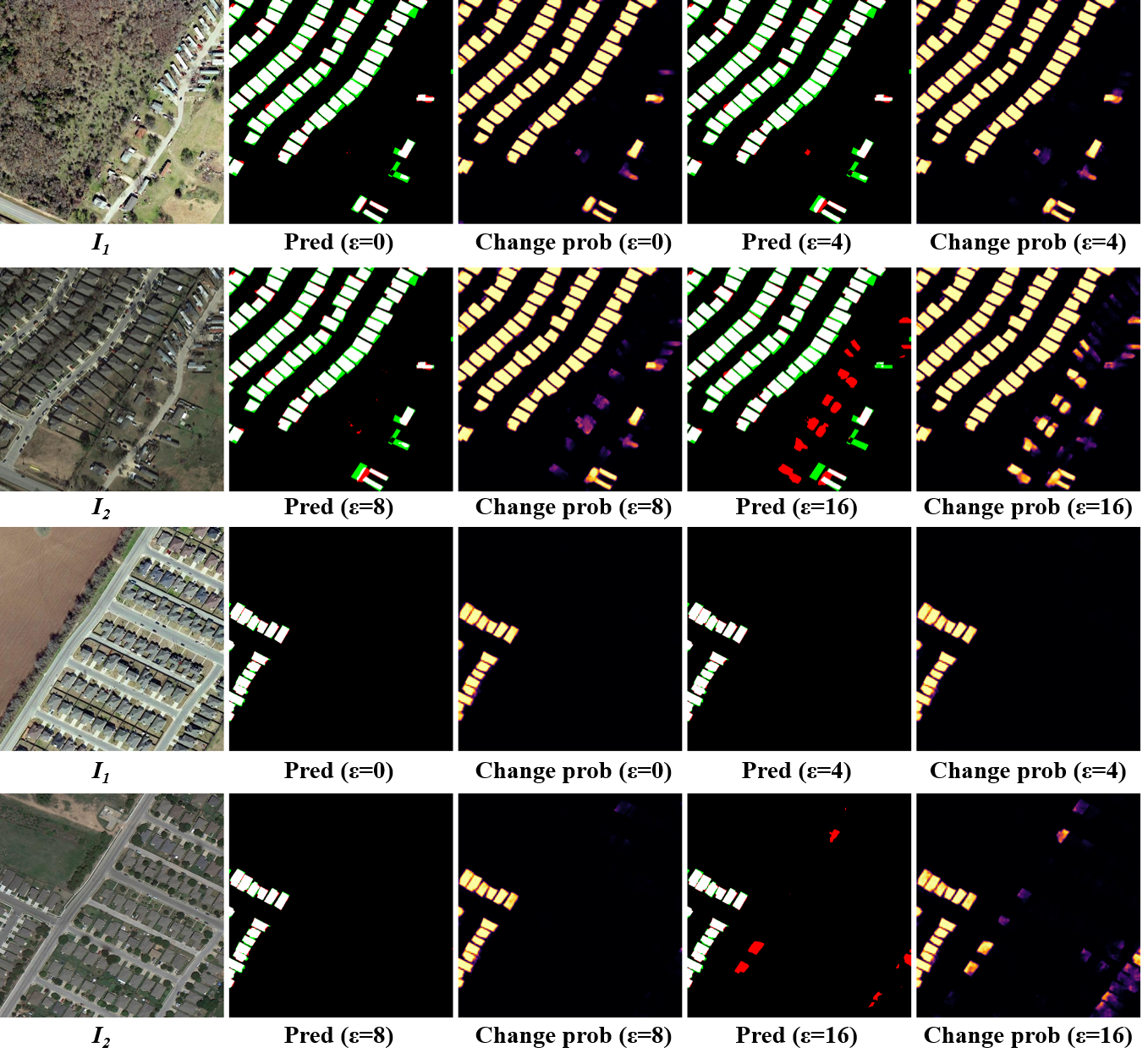}
    \caption{ChessMamba CD predictions with spatial shifts $\epsilon = 4, 8, 16$ pixels. Green/red: FN/FP regions.}
\label{fig.vis_shift}
\end{figure}

\textbf{Robustness to Spatial Shift:} 
To assess the robustness of our ChessMamba model to spatial misalignments in bitemporal observations, we conduct a qualitative evaluation on the LEVIR-CD dataset. We apply controlled pixel shifts ($\epsilon = 4, 8, 16$ pixels) to the pre-change image ($T_1$), while keeping the post-change image ($I_2$) fixed as the reference. Predictions from shifted inputs are visualized in Fig. \ref{fig.vis_shift} through comparative error maps (TP/FP/FN) and probability heatmaps.

The results demonstrate that small shifts (e.g., 4 and 8 pixels) introduce minimal degradation, with predictions maintaining high fidelity and low false positives/negatives. However, as shifts further increase (16 pixels), error maps reveal growing FN regions,. This highlights ChessMamba exhibits resilience to minor misalignments, validating the efficacy of its MCA-SSM module in handling subtle spatial variances.

\begin{figure}[h]
\centering
    \includegraphics[width=1\linewidth]{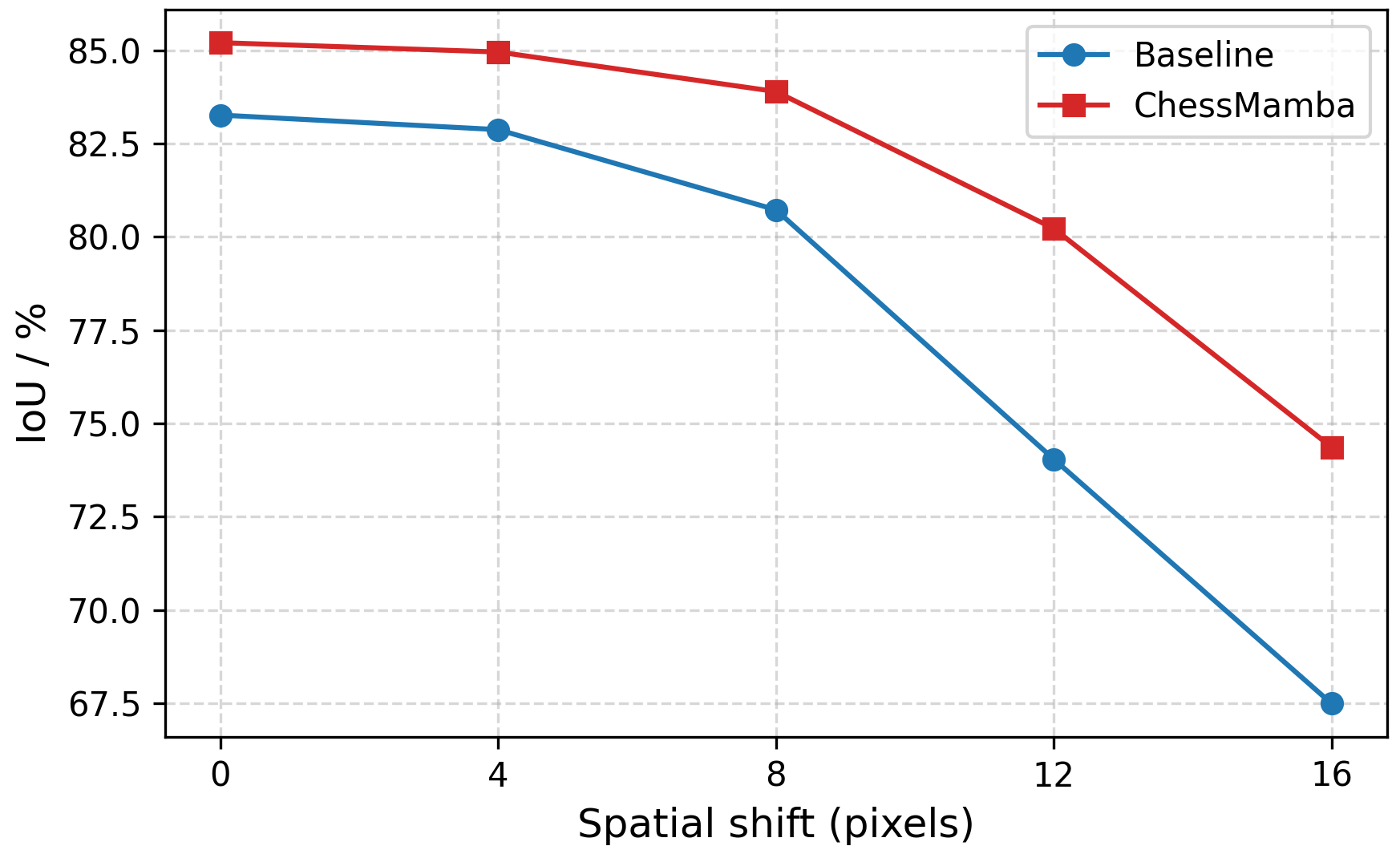}
    \caption{IoU of ChessMamba and baseline under different spatial shifts.}
\label{fig.shift_curve}
\end{figure}

Figure \ref{fig.shift_curve} further presents the IoU performance of ChessMamba versus the baseline across increasing spatial shifts. ChessMamba consistently outperforms the baseline, maintaining higher IoU with minimal degradation. At 0-pixel shift, ChessMamba holds a $\sim$2\% advantage. As the shift increases to 16 pixels, this performance gap widens significantly. ChessMamba has a less 10.84\% IoU drop compared to the baseline's drop of 15.77\%. These results demonstrate ChessMamba has superior robustness to spatial misalignment.

\textbf{Alignment of Heterogeneous Features:}
We qualitatively assess the heterogeneous feature alignment of our ChessMamba-based model on the BRIGHT dataset. Fig.\ref{fig.vis_shift} provides paired pre-disaster optical (RGB) and post-disaster SAR imagery with building-level damage labels. At three spatial scales of the MCA-SSM module, we measure the discrepancy between RGB and SAR streams as the per-location $L_2$ distance, yielding "befor" and "after" alignment maps.

These results show clear qualitative improvements in cross-modal consistency after alignment. Before applying MCA-SSM, discrepancy maps are dominated by artifacts such as SAR clutter and background texture, with high responses scattered across both damaged and intact regions. After alignment, responses become more structured and concentrated along building footprints and damaged areas, while non-building regions (roads, open land, noise) are strongly suppressed. Across three feature scales, the post-alignment maps highlight semantically meaningful structures rather than raw sensor patterns, indicating that the module effectively reduces modality bias and promotes a shared geometric representation aligning with the damage labels.

\begin{figure}[t]
\centering
    \includegraphics[width=1\linewidth]{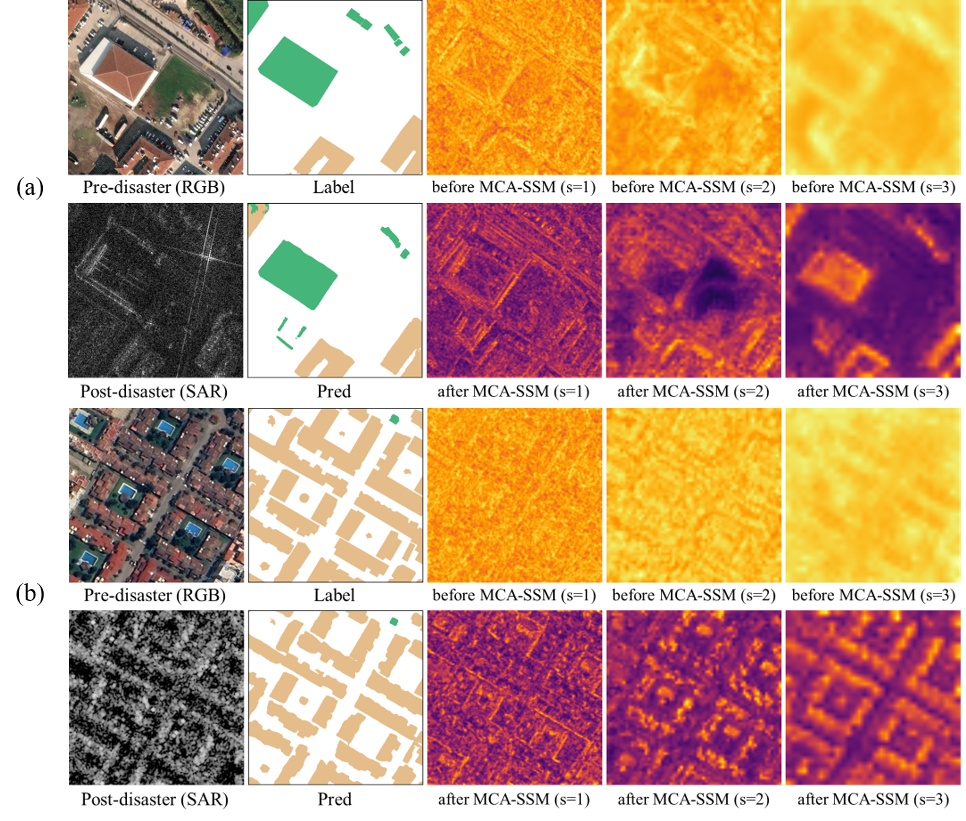}
    \caption{Visualization of the features before and after applying MCA-SSM. The damage scenes are: (a) earthquake, (b) hurricane. }
\label{fig.vis_feat}
\end{figure}

We further employ t-SNE to analyze semantic representations from BRIGHT test features. For each annotated building, we sample a feature vector at 1/16 resolution from the MCA-SSM stage, separately extracting pre-disaster optical and post-disaster SAR embedidngs. Using balanced samples across damage categories (Intact/Damaged/Destroyed), we compare t-SNE projections before and after MCA-SSM.

Before applying MCA-SSM, features form a modality-separated manifold with severe category overlap, revealing dominant sensor-specific bias over semantic discriminability. After using MCA-SSM, RGB features collapses to a tight cluster. This indicates homogeneity in pre-disaster appearances with suppressed sensor/noise variations. Conversely, SAR features separate into compact clusters reflecting different disaster types in the dataset (e.g., earthquake, hurricane, flood), confirming semantically structured reparameterization. This enhanced intra-class compactness and category separation demonstrate MCA-SSM’s efficacy in aligning cross-modal features for optimized change classification.

\subsection{Comparative Experiments}


\textbf{Binary Change Detection:}  Table \ref{Table.Levir} compares ChessMamba against recent state-of-the-art methods on Levir-CD, including CNN-based (A2Net \cite{li2023lightweight}, MDIPNet \cite{chang2024remote}, SEIFNet \cite{huang2024spatiotemporal}, STADE-CDNet \cite{li2024stade}), Transformer-driven (ChangeFormer \cite{bandara2022transformer}), self-attention designs \cite{dong2025relation} and Mamba variants (ChangeMamba \cite{chen2024changemamba}). Specifically, while RHighNet’s relation modeling yield high Precision (93.97\%), its Recall remains constrained. SAM-CD \cite{ding2024adapting}, leveraging pre-trained foundation model, also obtains competitive results.

\begin{figure}[t]
\centering
    \includegraphics[width=1\linewidth]{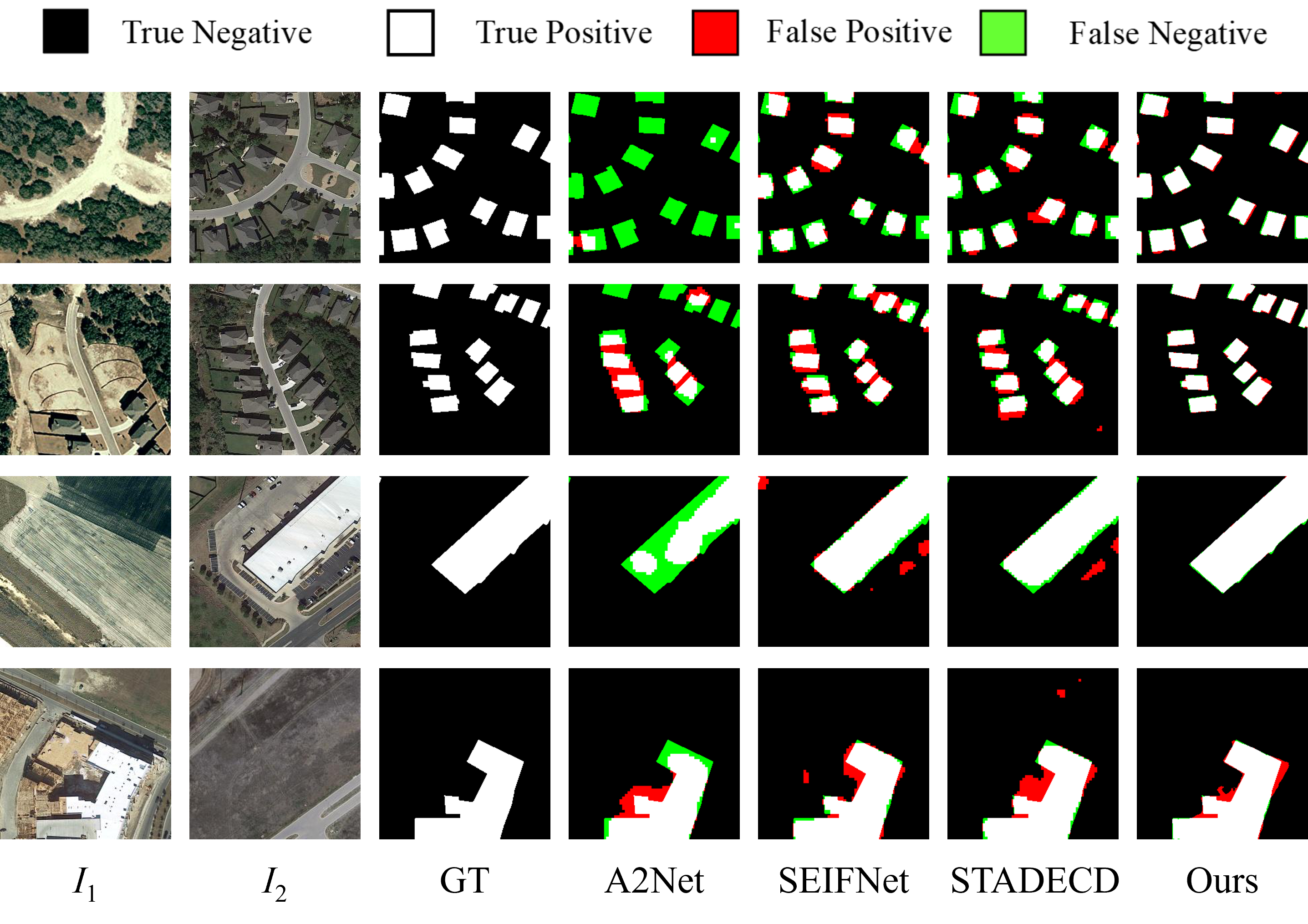}
    \caption{Qualitative results of different BCD methods (Levir).}
\label{fig.bcd_compare}
\end{figure}

ChessMamba achieves superior IoU (85.20\%), outperforming all alternatives. It attains the highest recall (90.69\%) while sustaining competitive precision (93.37\%), establishing a balanced precision-recall profile essential for reliable change mapping. This stems from topologically coherent state encoding that preserves neighborhood integrity during fusion. As illustrated in Fig. \ref{fig.bcd_compare}, this paradigm significantly reduces omission errors in fragmented structures while suppressing registration-induced false positives (+0.94\% IoU over the SOTA), validating the efficacy of structured temporal interaction.

\begin{table}[t]
\centering
\caption{Qualitative results of different BCD methods (Levir).}
\resizebox{1\linewidth}{!}{%
\begin{tabular}{l|c|ccccc}
\toprule
\rowcolor[HTML]{EFEFEF} 
Methods & \small Reference & OA & $Pre$ & $Rec$ & $F_1$ & IoU \\
\midrule
ChangeFormer \cite{bandara2022transformer} & \small IGARSS'22 & 99.04 & 92.05 & 88.80 & 90.40 & 82.48 \\
A2Net \cite{li2023lightweight} & \small TGRS'23 & 98.95 & 92.96 & 85.81 & 89.24 & 80.73 \\
DMINet \cite{feng2023change} & \small TGRS'23 & 99.07 &92.52 & 89.95 & 90.71 & 82.99 \\
SAM-CD \cite{ding2024adapting} & \small TGRS'24 & 99.14 & 92.23 & 90.68 & 91.46 & 84.26 \\
MDIPNet \cite{chang2024remote} & \small TIP'24 & 99.14 & 92.04 & 90.22 & 91.12 & 83.69 \\
SEIFNet \cite{huang2024spatiotemporal} & \small TGRS'24 & 99.09 & 92.49 & 89.46 & 90.95 & 83.40 \\
STADE-CDNet \cite{li2024stade} & \small TGRS'24 & 97.32 & 87.61 & 83.25 & 85.27 & 76.71 \\
ChangeMamba \cite{chen2024changemamba} & \small TGRS'24 & 99.04 & 90.73 & 90.30 & 90.51 & 82.67 \\
DDPM-CD \cite{bandara2025ddpm} & \small WACV'25 & 99.09 & 92.45 & 89.41 & 90.91 & 83.35 \\
RHighNet \cite{dong2025relation} & \small TGRS'25 & \textbf{99.32} & \textbf{93.97} & 88.55 & 91.18 & 83.79 \\
\midrule
ChessMamba & \small \textit{proposed} & 99.20 & 93.37 & \textbf{90.69} & \textbf{92.01} & \textbf{85.20} \\
\bottomrule
\end{tabular}}
\label{Table.Levir}
\end{table}

\textbf{Building Damage Assessment:} On the multimodal BRIGHT dataset, ChessMamba achieves new SOTA performance (68.40\% mIoU), surpassing recent methods including disaster-specialized Transformers (DamageFormer \cite{chen2022dual}: 66.09\% mIoU), graph-enhanced state space models (GSTM-SCD \cite{liu2025gstm}: 66.83\% mIoU), and foundation-based approaches (ChangeOS \cite{zheng2021building}: 65.98\% mIoU). DamageFormer’s attention fusion and GSTM-SCD’s spatiotemporal reasoning yield moderate damage classification ($F_1^{clf}$: ~74\%) but incur localization instability ($F_1^{loc}$ 90.51\%).

Our framework attains leading localization accuracy (90.92\% $F_1^{loc}$) while maintaining precise segmentation (71.42\% $F_1^{clf}$). This reflects its phase-aligned partitioned feature learning, which enforces cross-modal consistency under severe registration shifts inherent in optical-SAR disaster scenes. Notably, in the damage categories including "damaged" (IoU 42.47\%) and "destroyed" classes (IoU 57.50\%), our topology-coherent paradigm reduces structural ambiguity in collapsed buildings (\~1.4\% over GSTM-SCD), confirming efficacy for fine-grained damage assessment under heterogeneity.

\begin{table*}[t]
\centering
\caption{Quantitative results of different BDA methods (BRIGHT dataset).}
\resizebox{0.8\linewidth}{!}{%
\begin{tabular}{l|c|cccc|cccc}
\toprule
\multirow{2}{*}{Methods} & \multirow{2}{*}{\small Reference} & \multicolumn{4}{c|}{IoU per class} & \multicolumn{4}{c}{Metrics} \\
\cmidrule(lr){3-6} \cmidrule(lr){7-10}
 & & Background & Intact & Damaged & Destroyed & $F_1^{loc}$ & $F_1^{clf}$ & OA & mIoU \\
\midrule
SiamAttnUNet \cite{adriano2021learning} & \small ISPRS' 21 & 95.83 & 69.45 & 36.03 & 49.96 & 87.21 & 70.22 & 94.99 & 62.82 \\
SiamCRNN \cite{chen2019change} & \small TGRS' 20 & 96.32 & 71.45 & 35.06 & 50.67 & 88.77 & 68.71 & 95.42 & 63.37 \\
ChangeOS \cite{zheng2021building} & \small RSE' 21 & 96.54 & 73.85  &38.99 & 54.53 & 89.60 & 71.88 & 95.84 & 65.98 \\
DamageFormer \cite{chen2022dual} & \small IGARSS' 22 & 96.75 & 74.30 & 40.35 & 52.98 & 89.92 & 72.22 & 95.95 & 66.09 \\
HGINet \cite{yao2024hierarchical} & \small TIP' 24 & 95.18 & 65.76 & 7.76 & 34.08 & 83.89 & 30.58 & 94.17 & 50.70 \\
Sigma \cite{wan2025sigma} & \small WACV' 25 & 95.86 & 70.15 & 35.27 & 47.95 & 86.82 & 64.20 & 95.16 & 62.31 \\
GSTM-SCD \cite{liu2025gstm} & \small ISPRS'25 & 96.63 & 73.75 & 41.09 & 55.86 & 90.51 & \textbf{73.96} & 95.88 & 66.83\\
\midrule
ChessMamba & \small \textit{proposed} & \textbf{97.09} & \textbf{76.52} & \textbf{42.47} & \textbf{57.50} & \textbf{90.92} & 71.42 & \textbf{96.39} & \textbf{68.40} \\
\bottomrule
\end{tabular}}
\label{Table.bright}
\end{table*}

\begin{figure}[t]
\centering
    \includegraphics[width=1\linewidth]{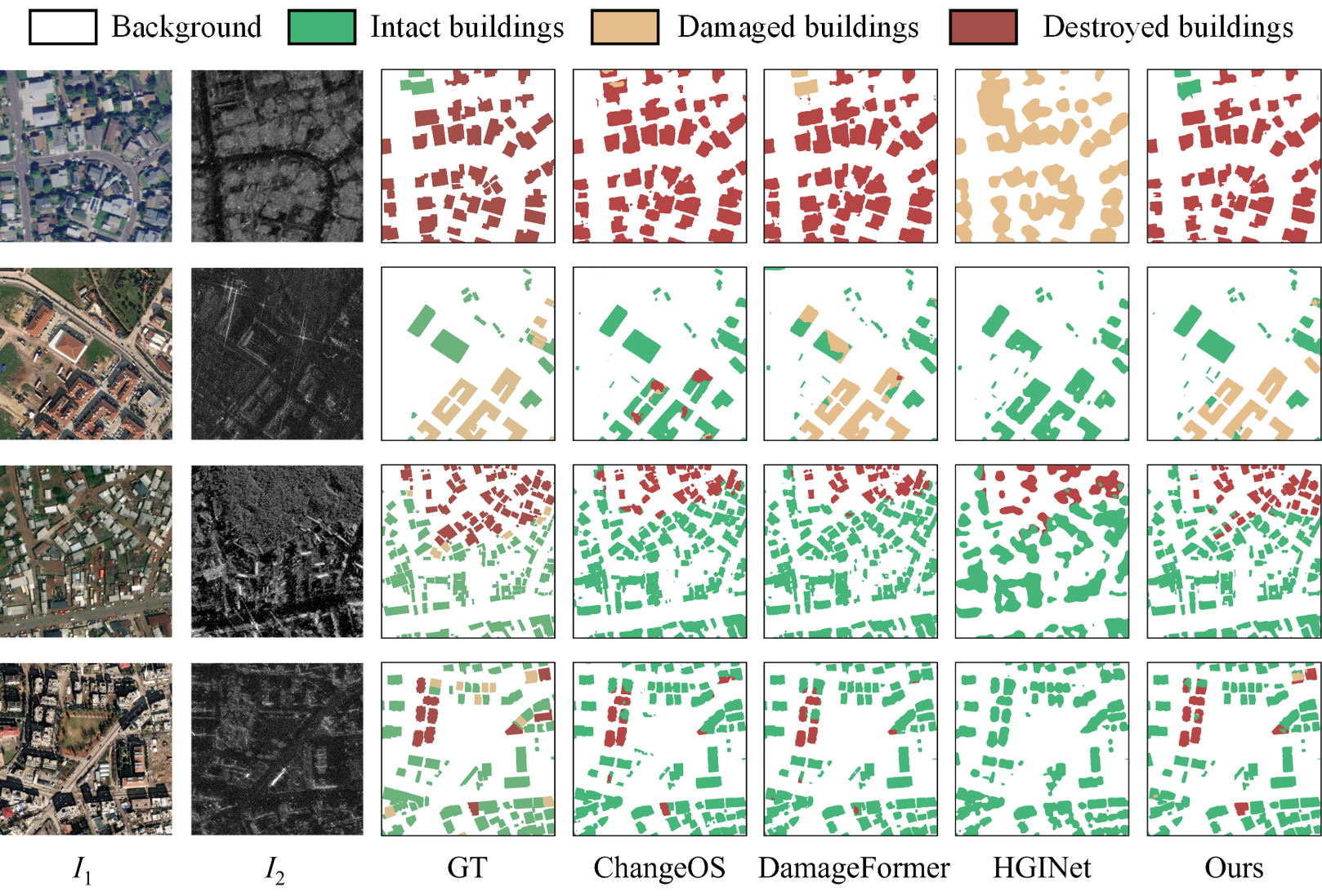}
    \caption{Qualitative results of different BDA methods (BRIGHT).}
\label{fig.bda_compare}
\end{figure}

\textbf{Semantic Change Detection:} On the SECOND dataset with 6 "from-to" land cover categories, ChessMamba achieves 65.32\% mIoU, establishing a new SOTA. t surpasses diverse competing frameworks: Bi-SRNet (73.41\% mIoU) with bi-temporal feature refinement, SCanNet (73.42\% mIoU) for spatiotemporal joint optimization, and the graph-enhanced SSM GSTM-SCD (73.50\% mIoU). While GSTM-SCD demonstrates strong relational reasoning, it exhibits semantic fragmentation at transition boundaries.

Our method has higher semantic precision and consistency, as reflected in the highest Sek of 24.80\%. This advantage stem from its core strength: preserving semantic continuity during bitemporal fusion through chessboard-constrained state propagation. This mechanism reduces semantic blurring in segmenting land cover transitions, particularly suppressing false transitions in seasonal variation zones. This elevates transition edge fidelity while efficiently modeling long-range dependencies. Critically, by maintaining geometric coherence during interleaved feature serialization, ChessMamba minimizes label conflict propagation across temporal states ($F_{scd}$ +0.97\%), a key limitation in prior sequence-based approaches.

\begin{table}[t]
\centering
\caption{Quantitative results of different SCD methods (SECOND).}
\resizebox{1\linewidth}{!}{%
\begin{tabular}{l|c|ccccc}
\toprule
\rowcolor[HTML]{EFEFEF} 
Methods & \small Reference & OA & mIoU & SeK & $F_{scd}$ \\
\midrule
Bi-SRNet \cite{ding2022bi} & \small TGRS'22 & 87.84 & 73.41 & 23.22 & 62.61 \\
Mamba-SCD \cite{chen2024changemamba} & \small TGRS'24 & 87.69 & 73.02 & 23.04 & 63.38 \\
SCanNet \cite{ding2024joint} & \small TGRS'24 & 87.86 & 73.42 & 23.94 & 63.66 \\
SAM-SCD \cite{mei2024scd} & \small TGRS'24 & 87.04 & 71.79 & 20.07 & 60.32 \\
M-CD \cite{paranjape2025mamba} & \small WACV'25 & 85.95 & 71.54 & 19.67 & 59.66 \\
GSTM-SCD \cite{liu2025gstm} & \small ISPRS'25 & 88.26 & 73.50 & 24.18 & 64.35 \\
\midrule
ChessMamba & \small \textit{proposed} & \textbf{90.08} & \textbf{73.62} & \textbf{24.80} & \textbf{65.32} \\
\bottomrule
\end{tabular}}
\label{Table.second}
\end{table}

\begin{figure*}[t]
\centering
    {\includegraphics[width=15cm]{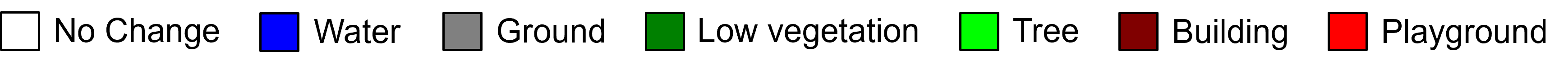}}\\
    \setlength{\tabcolsep}{1pt}
    \begin{tabular}{>{\centering\arraybackslash}m{0.6cm}>{\centering\arraybackslash}m{1.7cm}>{\centering\arraybackslash}m{1.7cm}>{\centering\arraybackslash}m{1.7cm}>{\centering\arraybackslash}m{1.7cm}>{\centering\arraybackslash}m{1.7cm}>{\centering\arraybackslash}m{1.7cm}>{\centering\arraybackslash}m{1.7cm}>{\centering\arraybackslash}m{1.7cm}>{\centering\arraybackslash}m{1.7cm}}
        (a1) &
        \includegraphics[width=1.6cm]{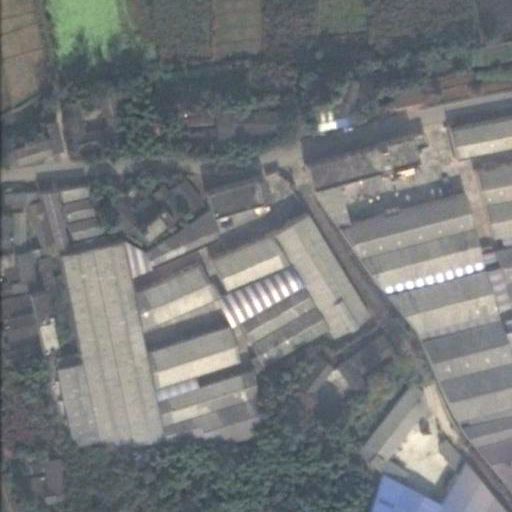} &
        \includegraphics[width=1.6cm]{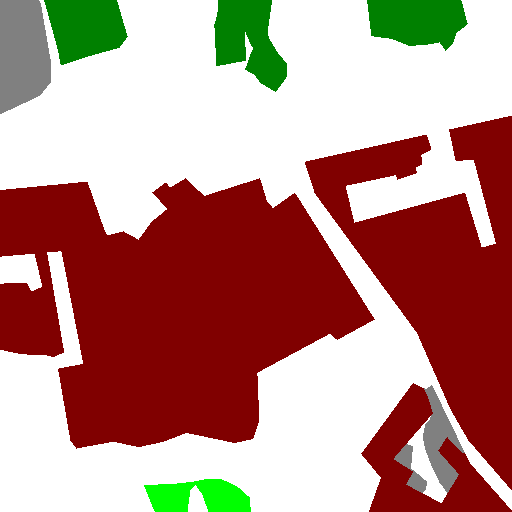} &
        \includegraphics[width=1.6cm]{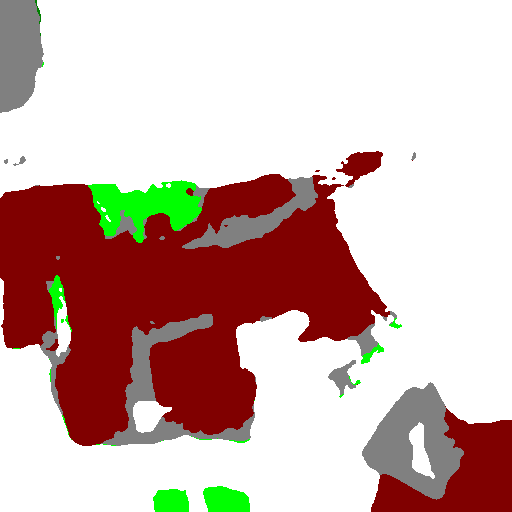} &
        \includegraphics[width=1.6cm]{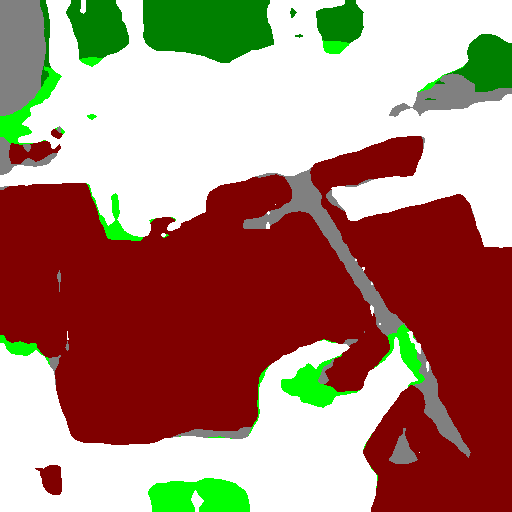} &
        \includegraphics[width=1.6cm]{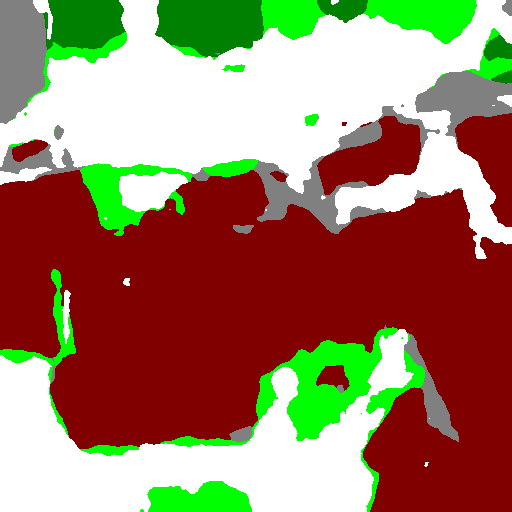} &
        \includegraphics[width=1.6cm]{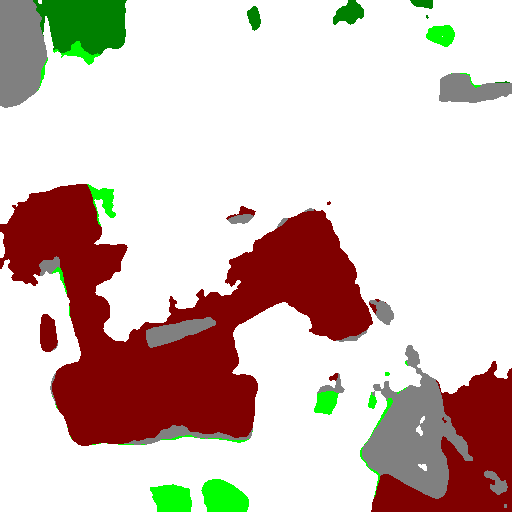} &
        \includegraphics[width=1.6cm]{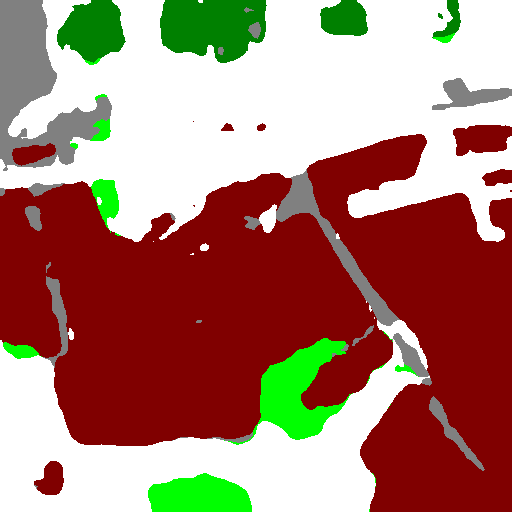} &
        \includegraphics[width=1.6cm]{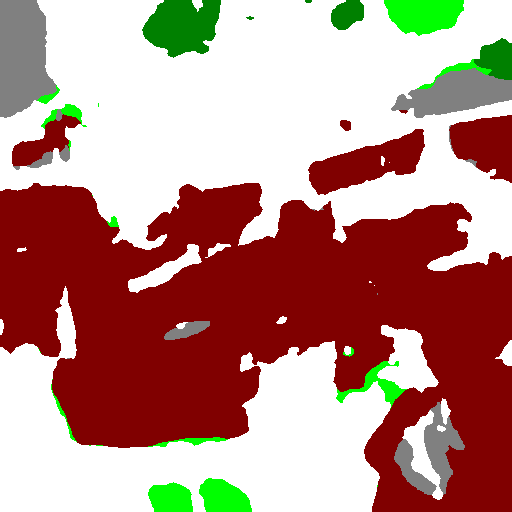} &
        \includegraphics[width=1.6cm]{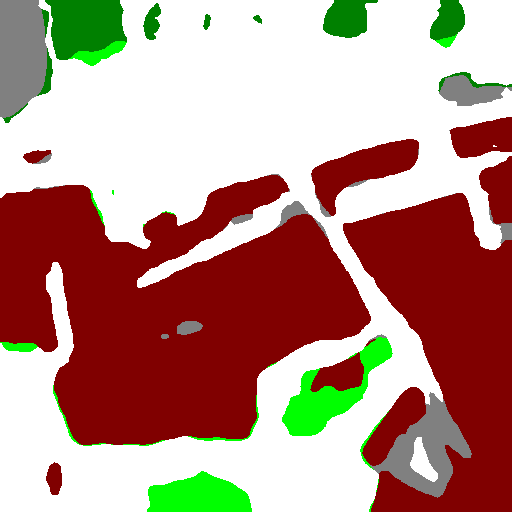} \\
        (a2) &
        \includegraphics[width=1.6cm]{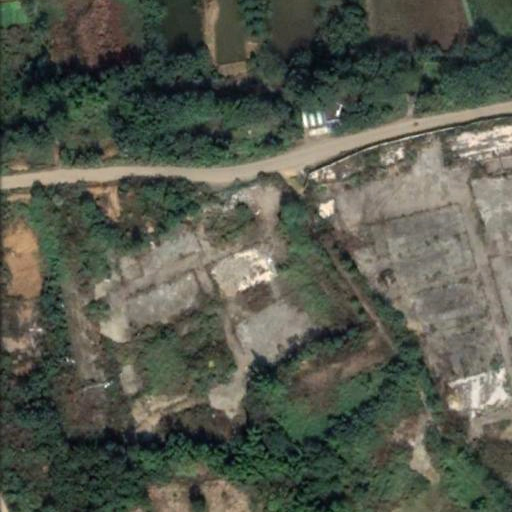} &
        \includegraphics[width=1.6cm]{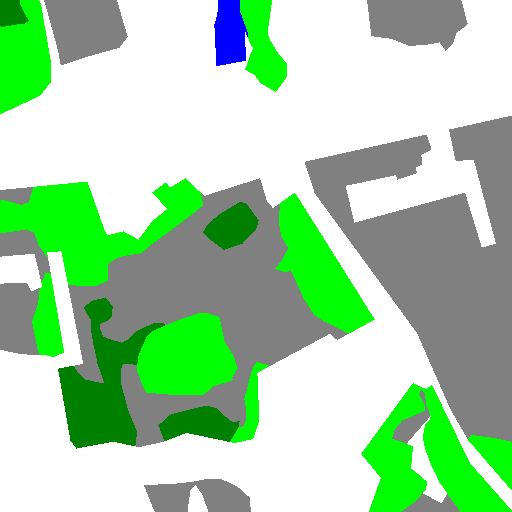} &
        \includegraphics[width=1.6cm]{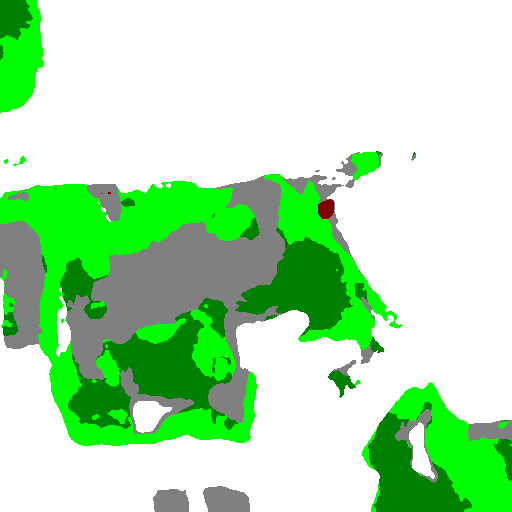} &
        \includegraphics[width=1.6cm]{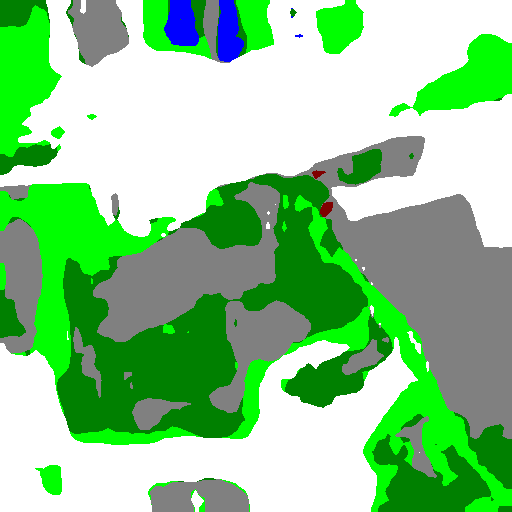} &
        \includegraphics[width=1.6cm]{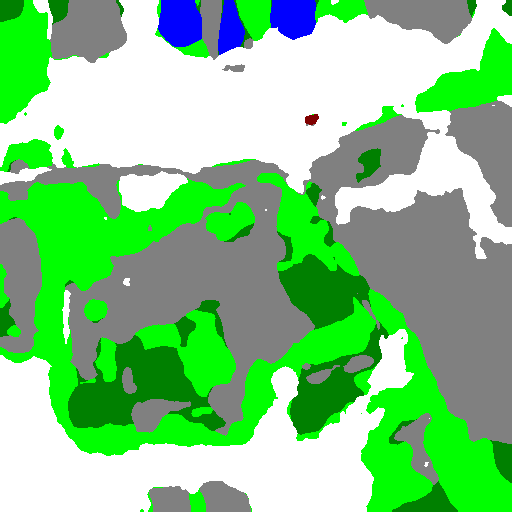} &
        \includegraphics[width=1.6cm]{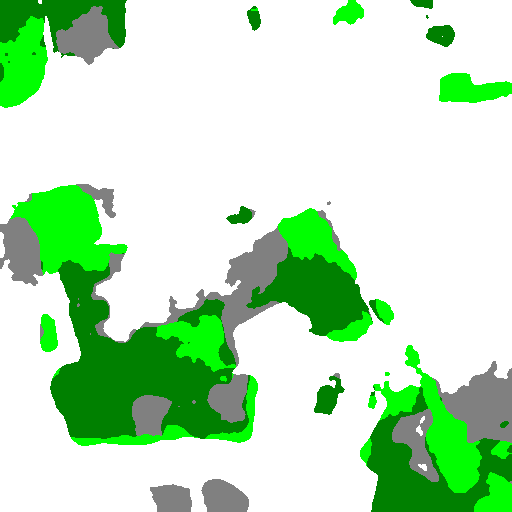} &
        \includegraphics[width=1.6cm]{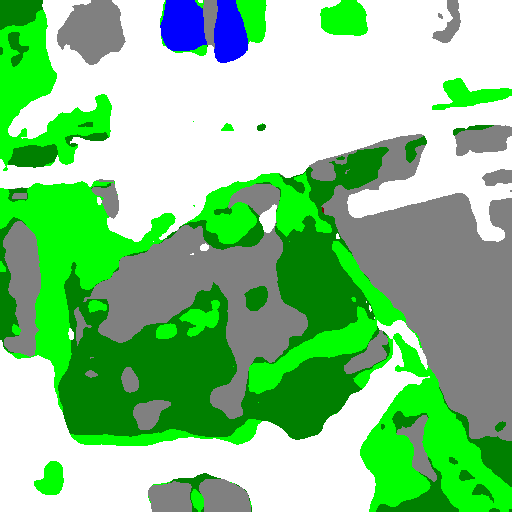} &
        \includegraphics[width=1.6cm]{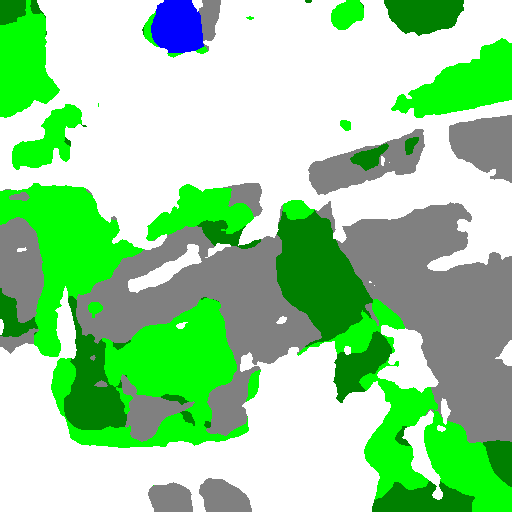} &
        \includegraphics[width=1.6cm]{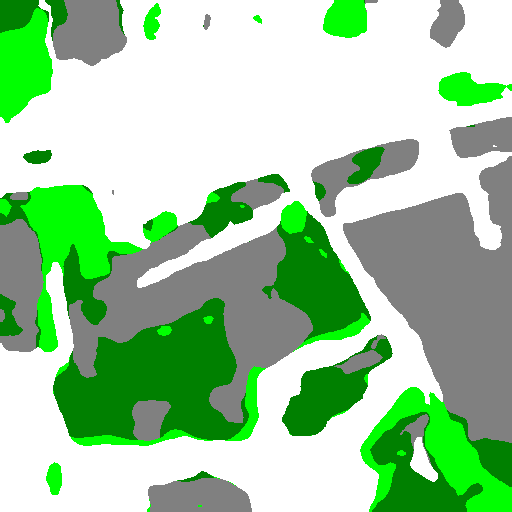} \\
        \hline\\
        (b1) &
        \includegraphics[width=1.6cm]{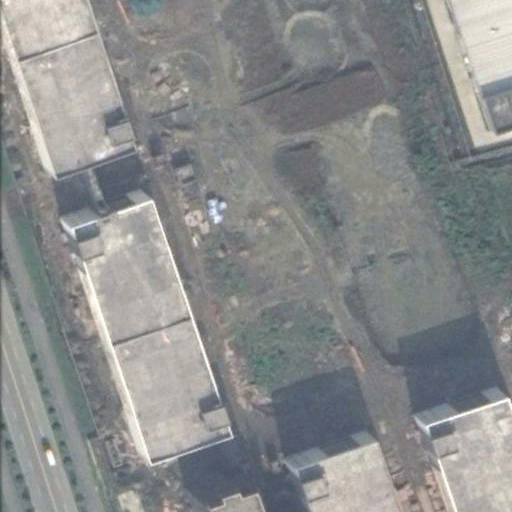} &
        \includegraphics[width=1.6cm]{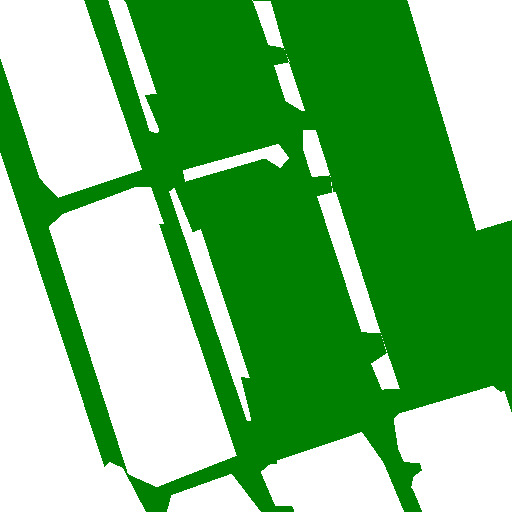} &
        \includegraphics[width=1.6cm]{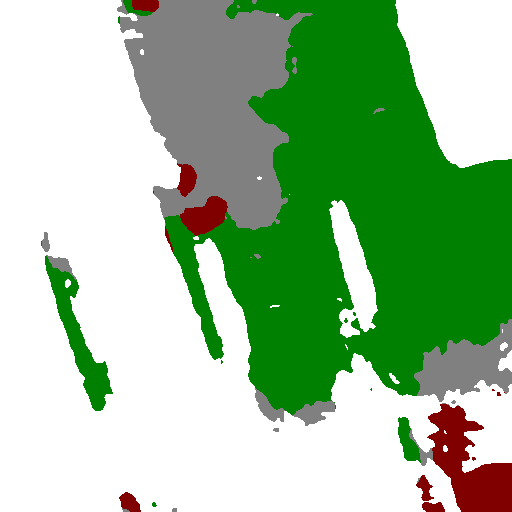} &
        \includegraphics[width=1.6cm]{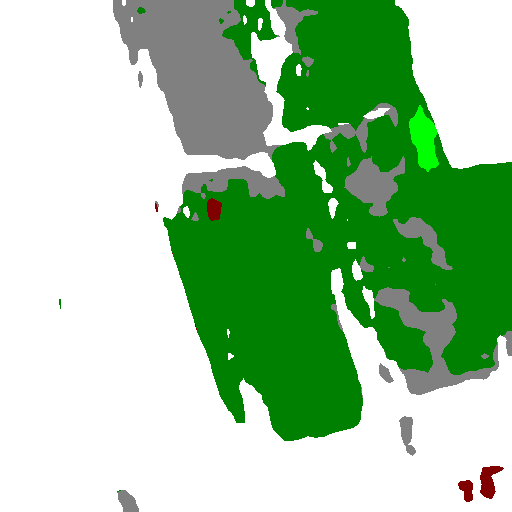} &
        \includegraphics[width=1.6cm]{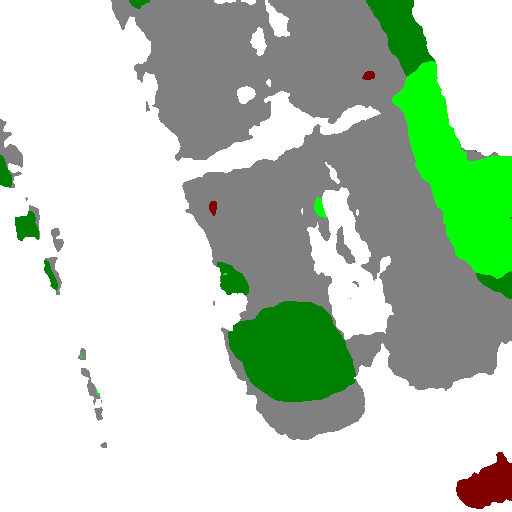} &
        \includegraphics[width=1.6cm]{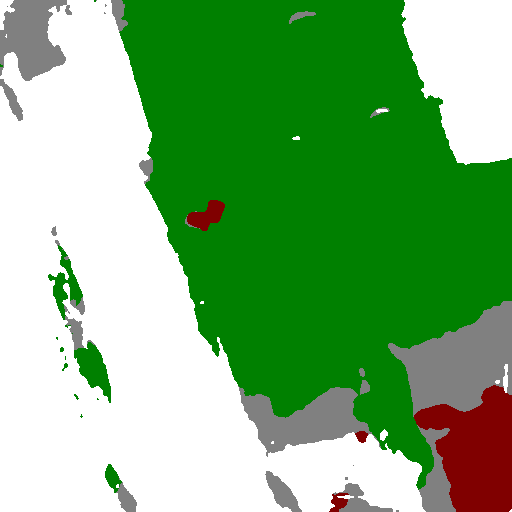} &
        \includegraphics[width=1.6cm]{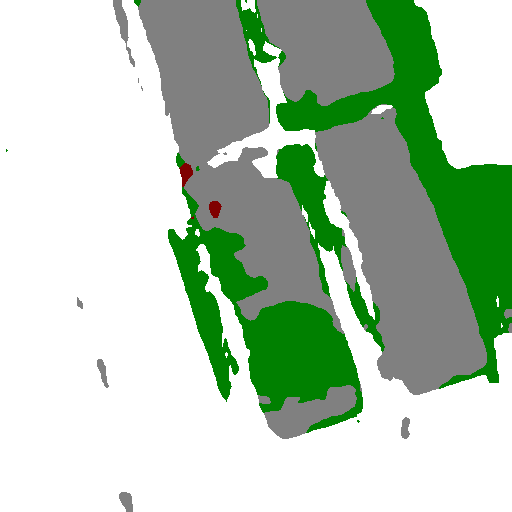} &
        \includegraphics[width=1.6cm]{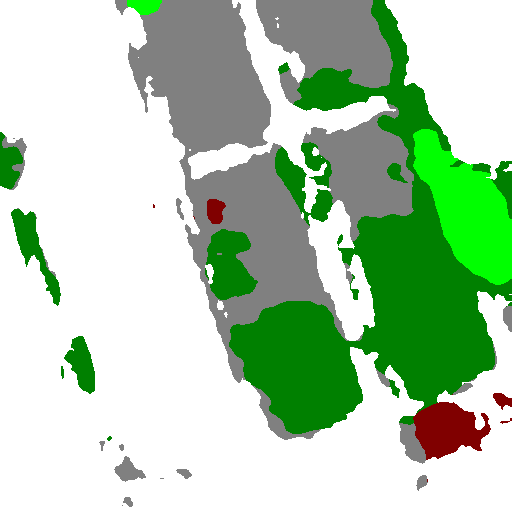} &
        \includegraphics[width=1.6cm]{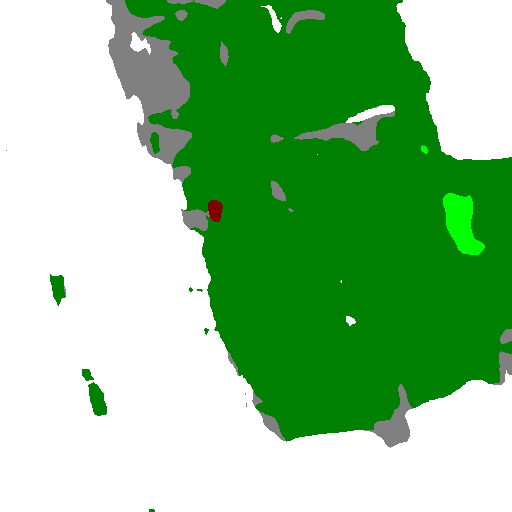} \\
        (b2) &
        \includegraphics[width=1.6cm]{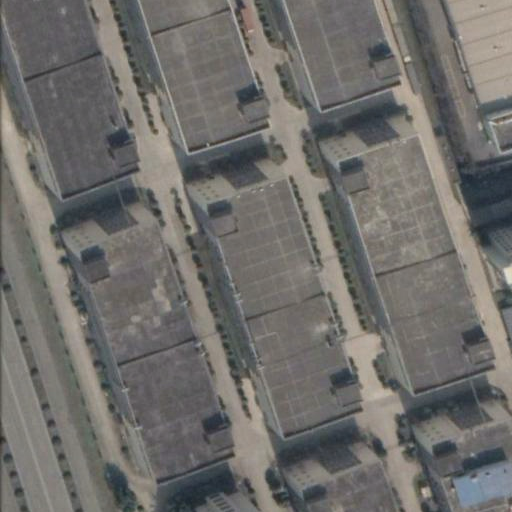} &
        \includegraphics[width=1.6cm]{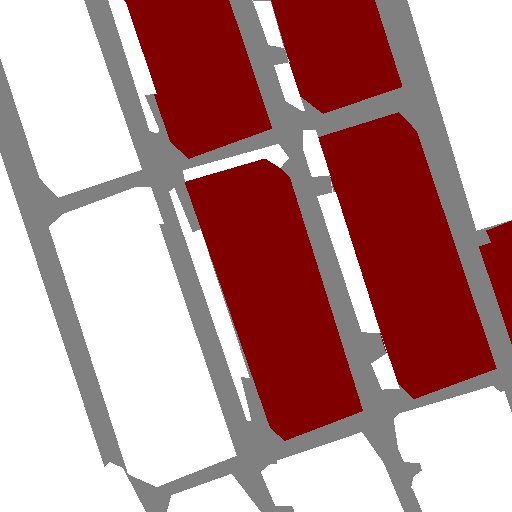} &
        \includegraphics[width=1.6cm]{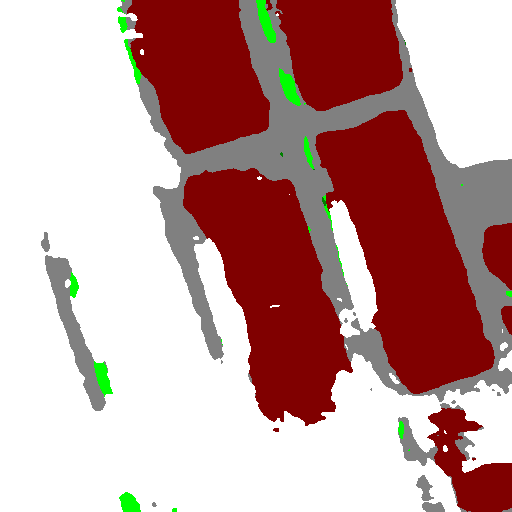} &
        \includegraphics[width=1.6cm]{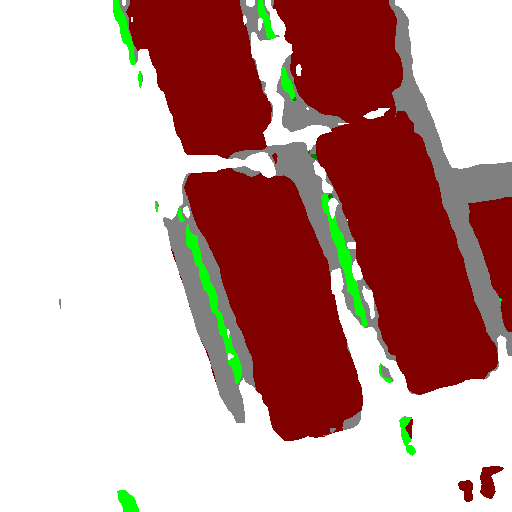} &
        \includegraphics[width=1.6cm]{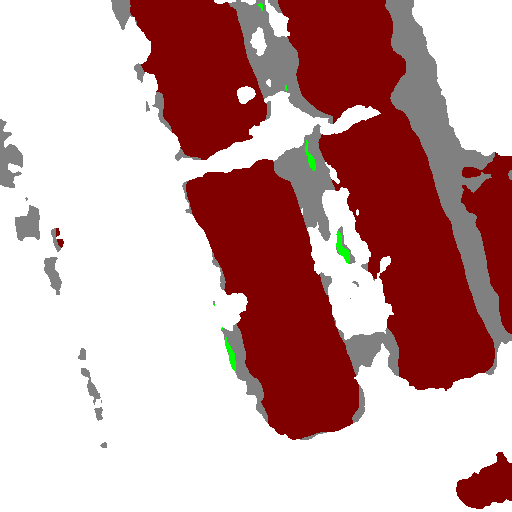} &
        \includegraphics[width=1.6cm]{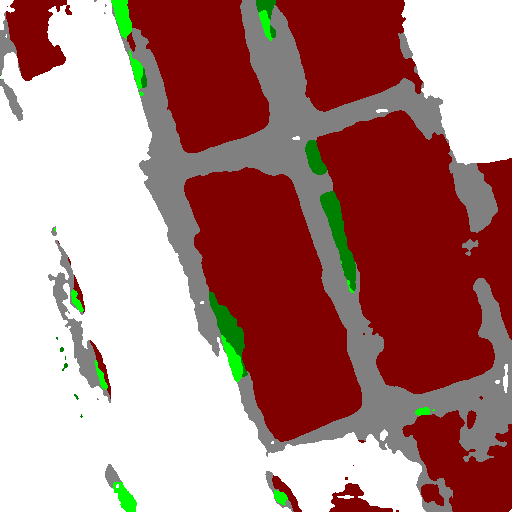} &
        \includegraphics[width=1.6cm]{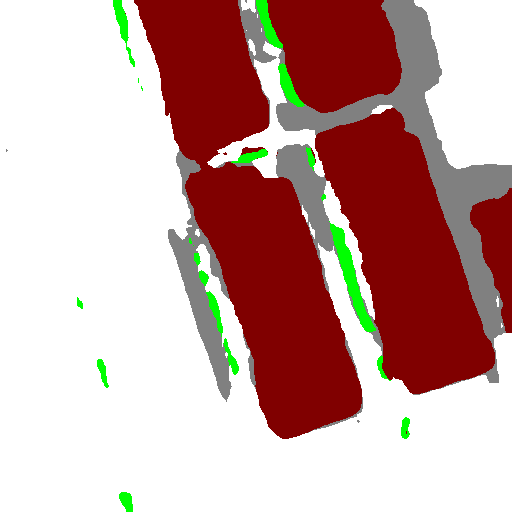} &
        \includegraphics[width=1.6cm]{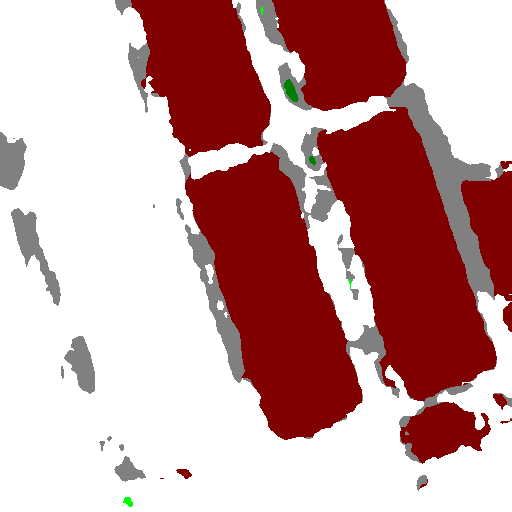} &
        \includegraphics[width=1.6cm]{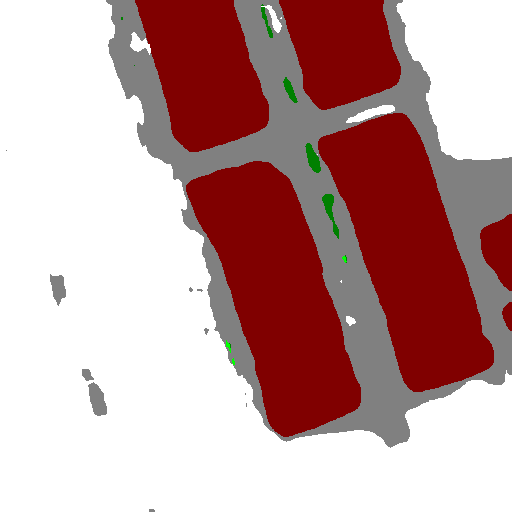} \\
        \hline\\
        (c1) &
        \includegraphics[width=1.6cm]{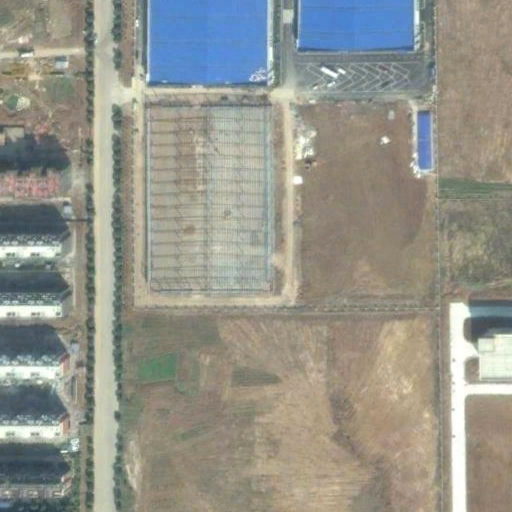} &
        \includegraphics[width=1.6cm]{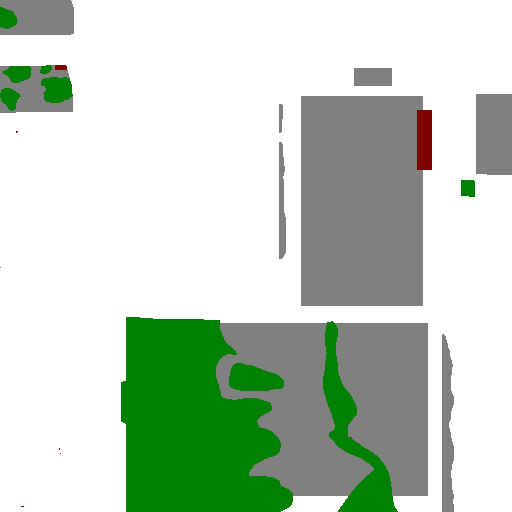} &
        \includegraphics[width=1.6cm]{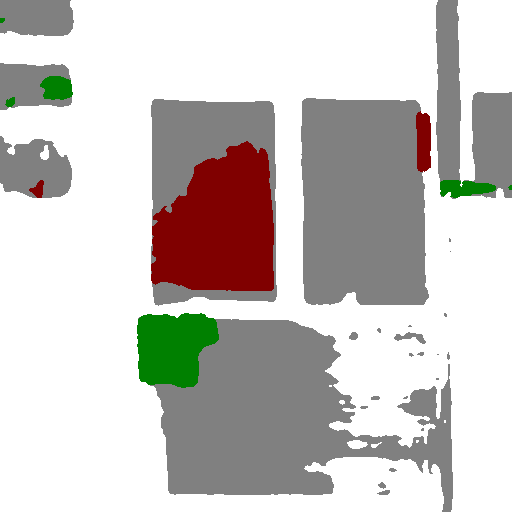} &
        \includegraphics[width=1.6cm]{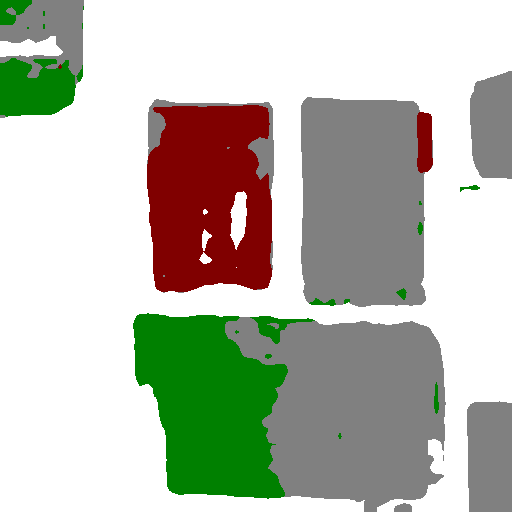} &
        \includegraphics[width=1.6cm]{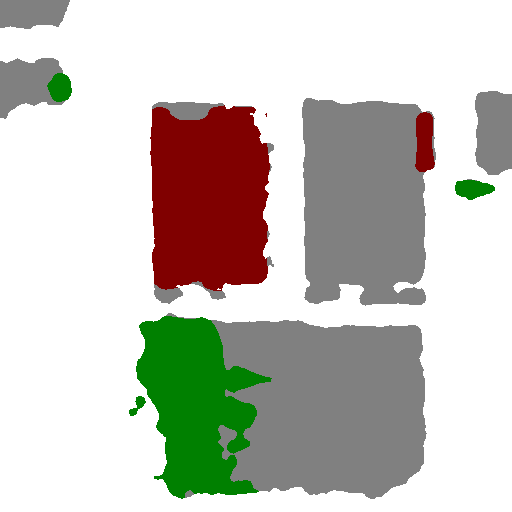} &
        \includegraphics[width=1.6cm]{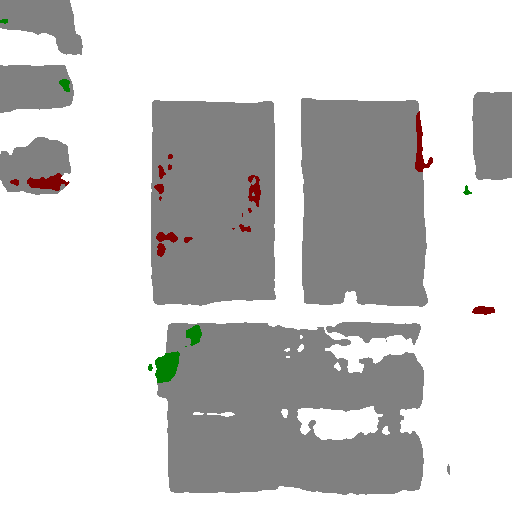} &
        \includegraphics[width=1.6cm]{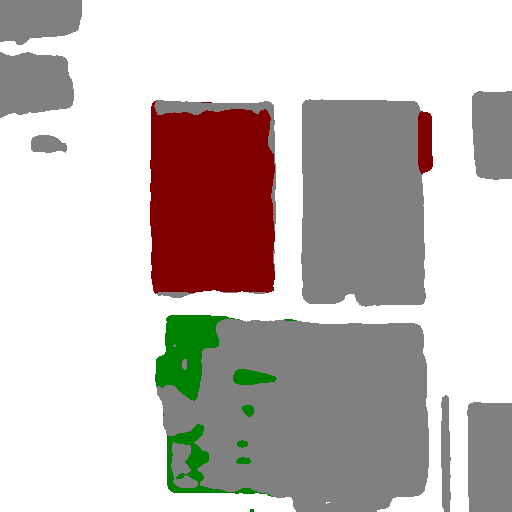} &
        \includegraphics[width=1.6cm]{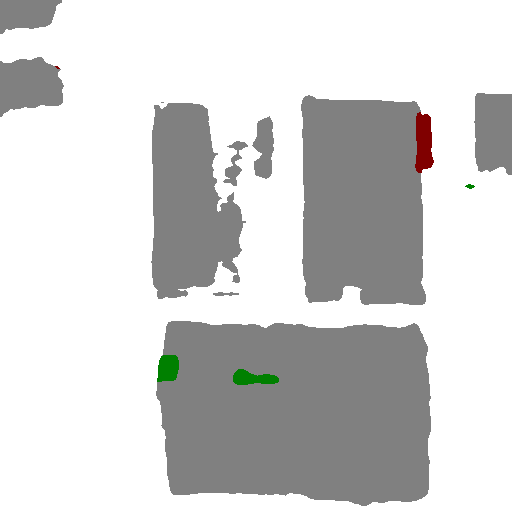} &
        \includegraphics[width=1.6cm]{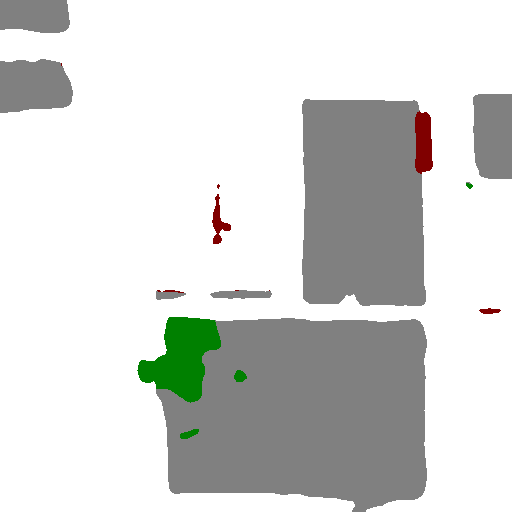} \\
        (c2) &
        \includegraphics[width=1.6cm]{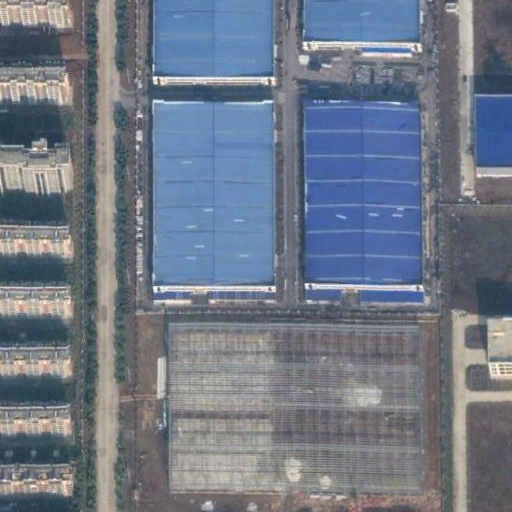} &
        \includegraphics[width=1.6cm]{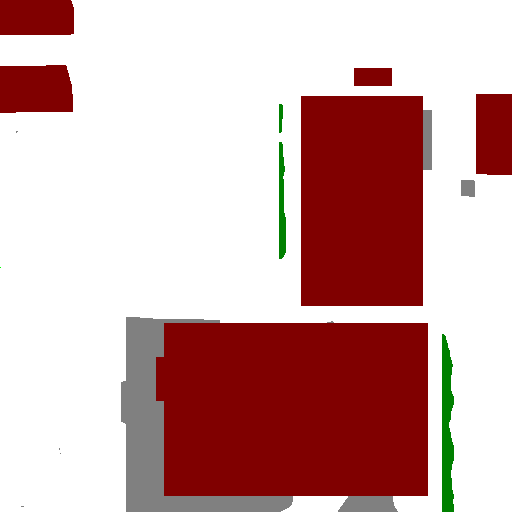} &
        \includegraphics[width=1.6cm]{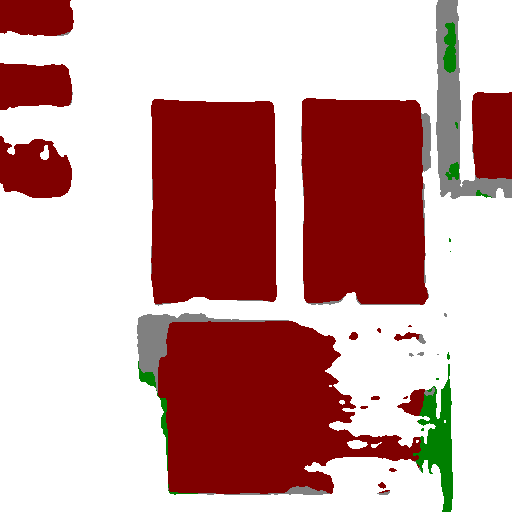} &
        \includegraphics[width=1.6cm]{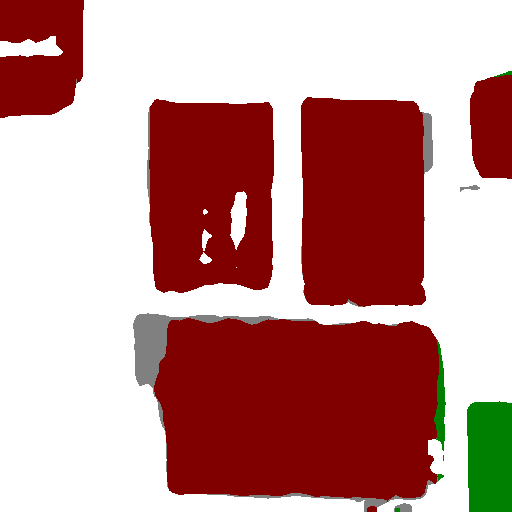} &
        \includegraphics[width=1.6cm]{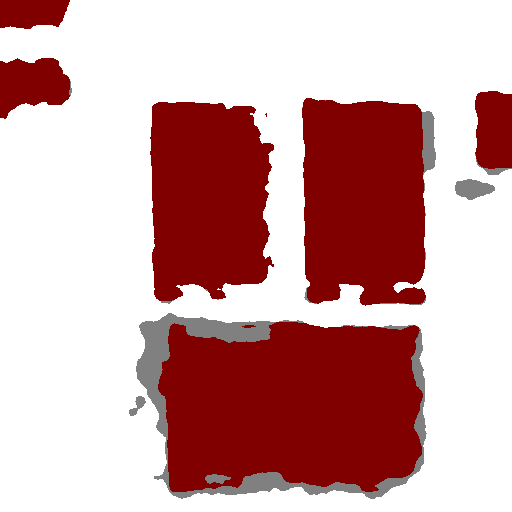} &
        \includegraphics[width=1.6cm]{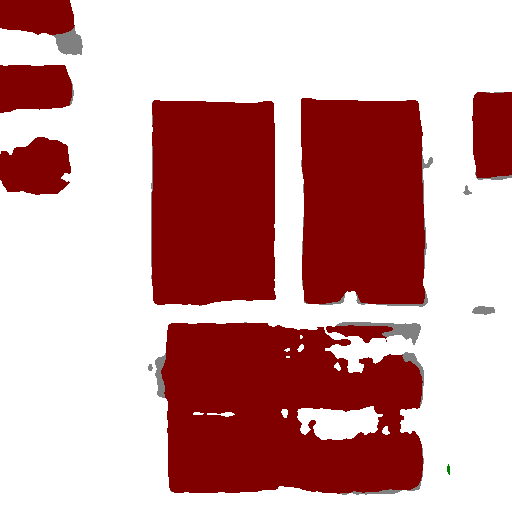} &
        \includegraphics[width=1.6cm]{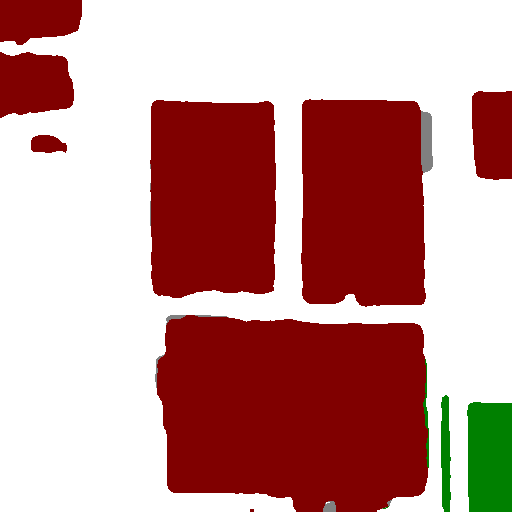} &
        \includegraphics[width=1.6cm]{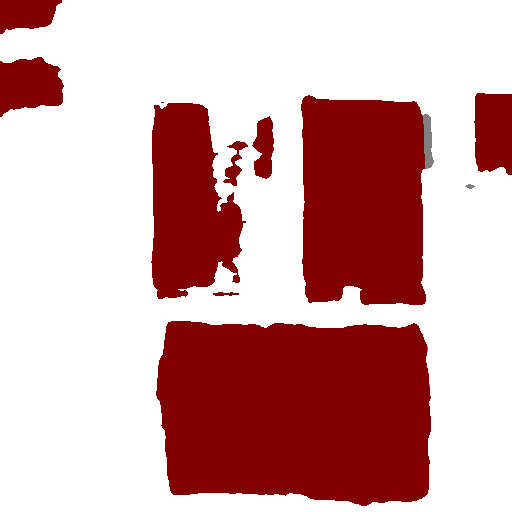} &
        \includegraphics[width=1.6cm]{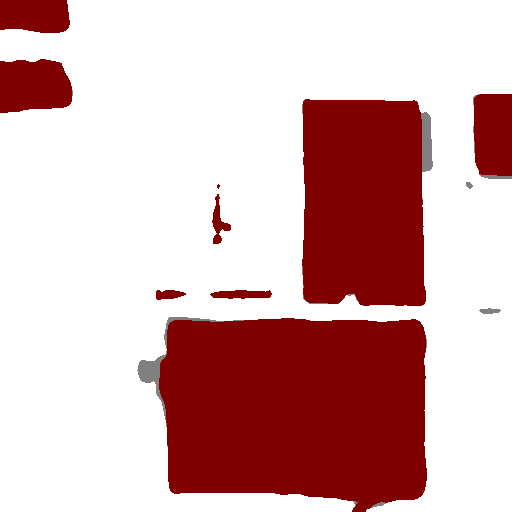} \\
        \hline\\
        (d1) &
        \includegraphics[width=1.6cm]{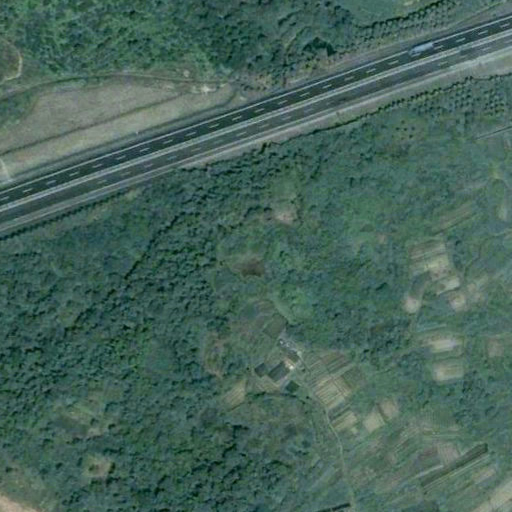} &
        \includegraphics[width=1.6cm]{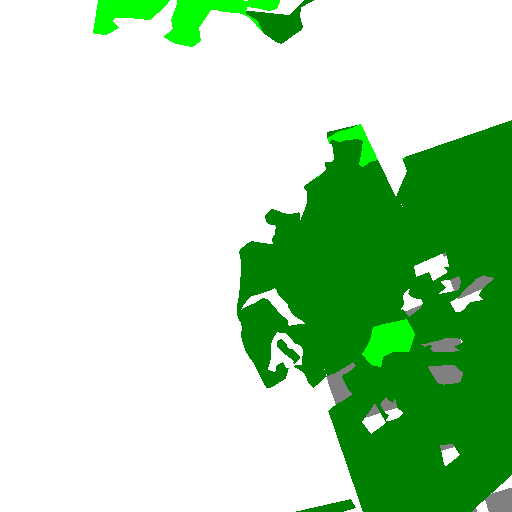} &
        \includegraphics[width=1.6cm]{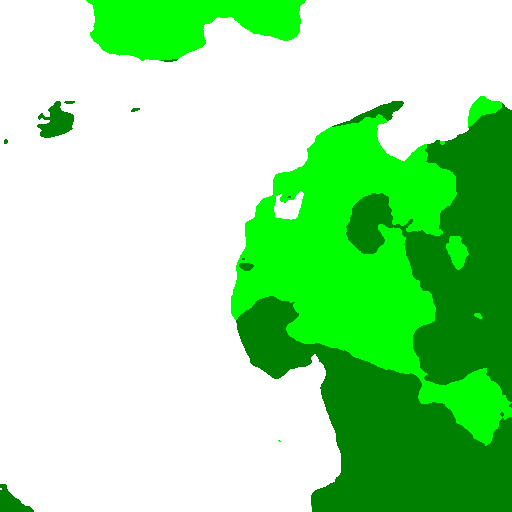} &
        \includegraphics[width=1.6cm]{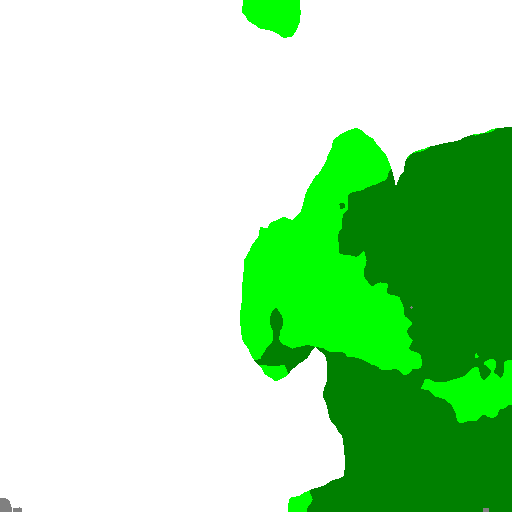} &
        \includegraphics[width=1.6cm]{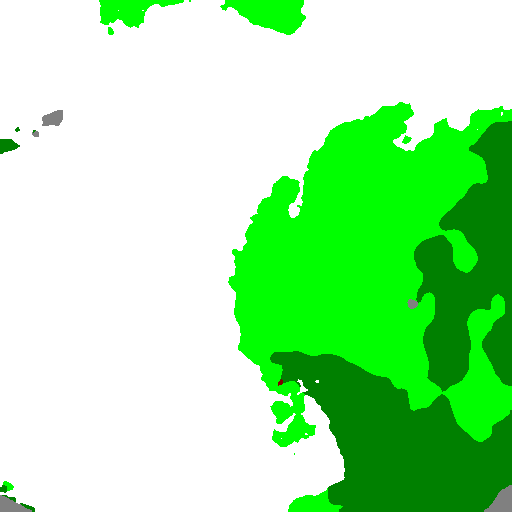} &
        \includegraphics[width=1.6cm]{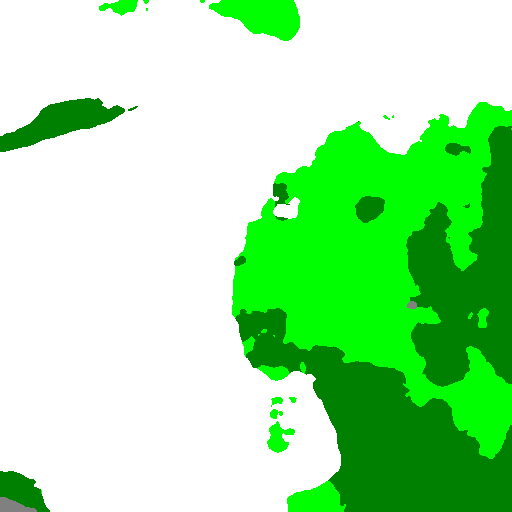} &
        \includegraphics[width=1.6cm]{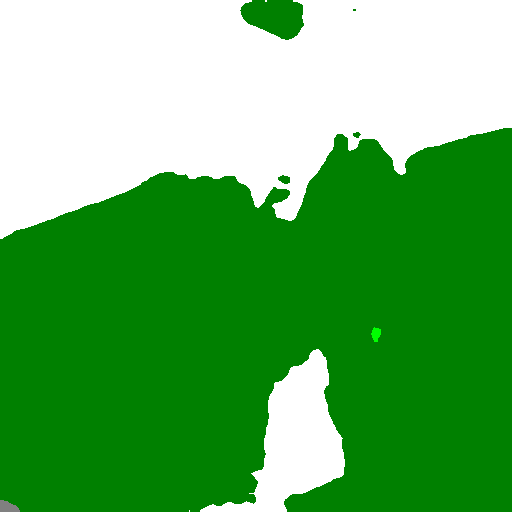} &
        \includegraphics[width=1.6cm]{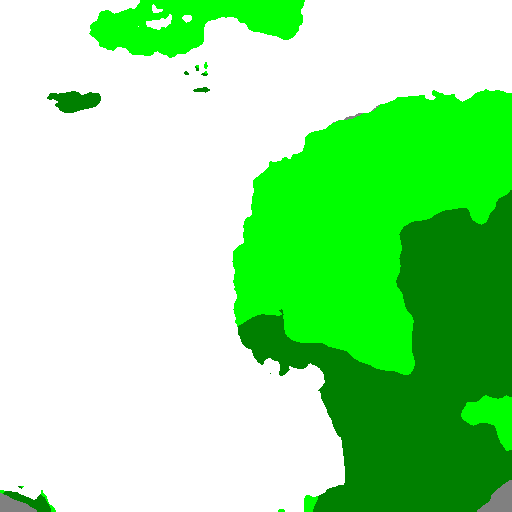} &
        \includegraphics[width=1.6cm]{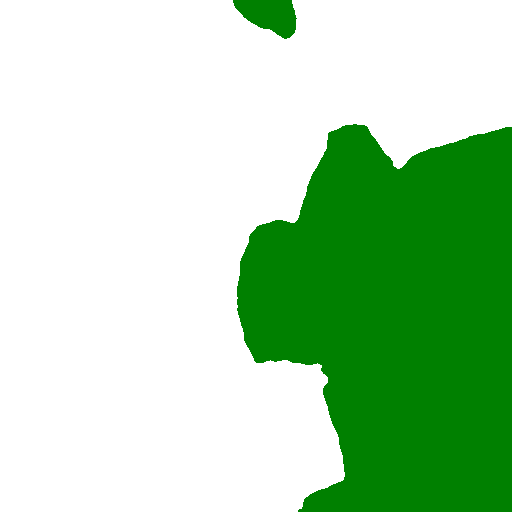} \\
        (d2) &
        \includegraphics[width=1.6cm]{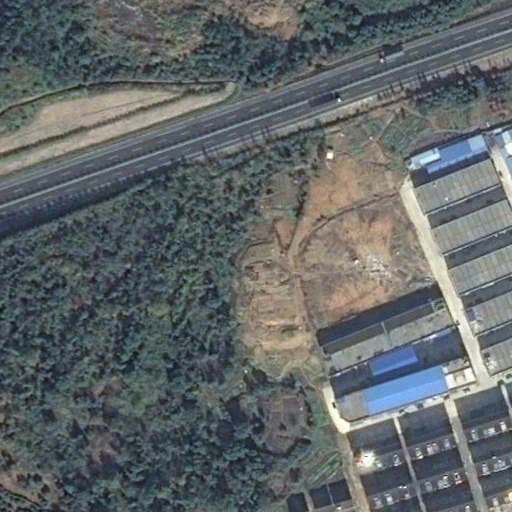} &
        \includegraphics[width=1.6cm]{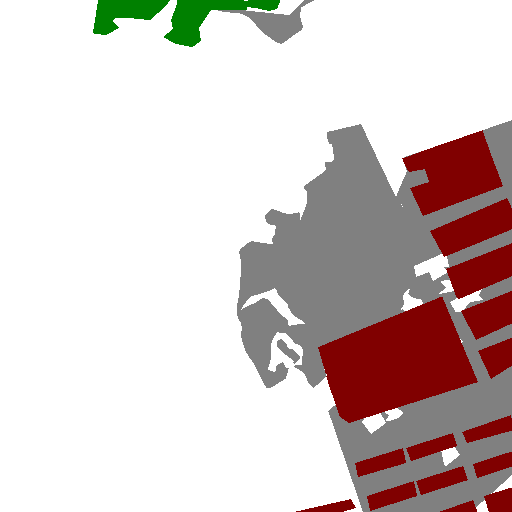} &
        \includegraphics[width=1.6cm]{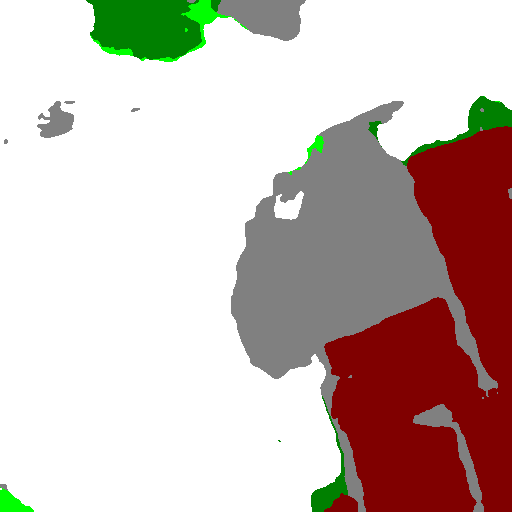} &
        \includegraphics[width=1.6cm]{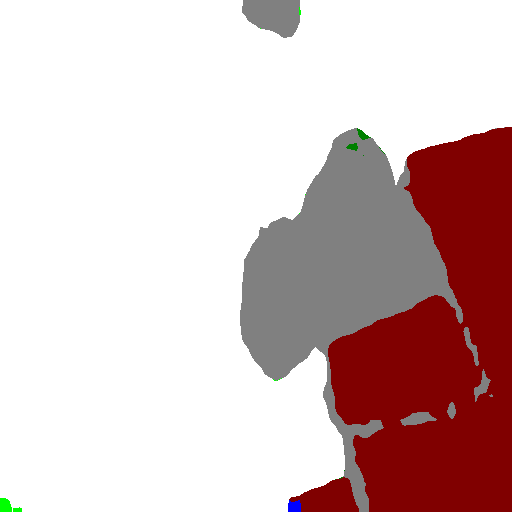} &
        \includegraphics[width=1.6cm]{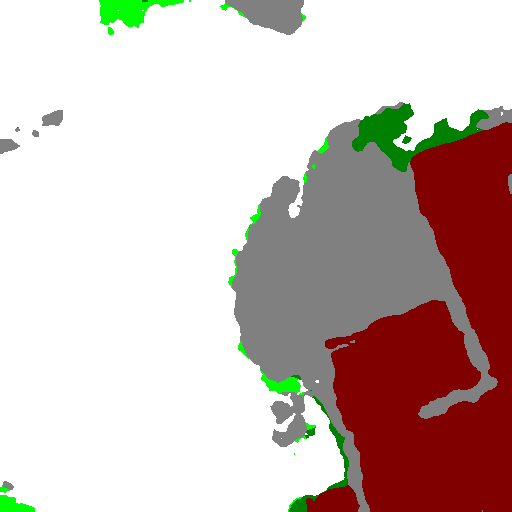} &
        \includegraphics[width=1.6cm]{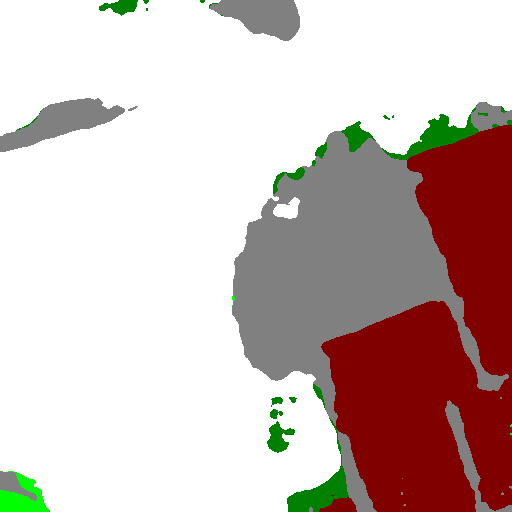} &
        \includegraphics[width=1.6cm]{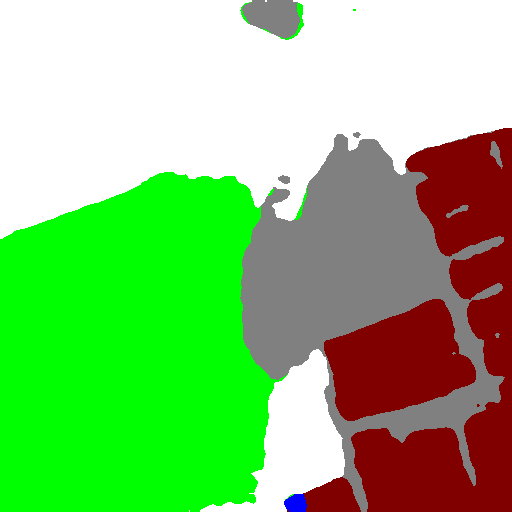} &
        \includegraphics[width=1.6cm]{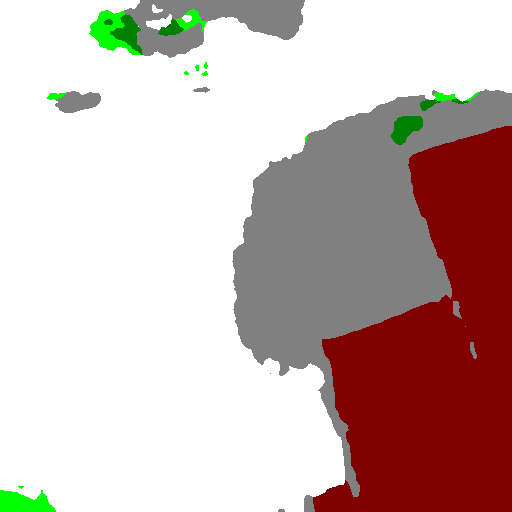} &
        \includegraphics[width=1.6cm]{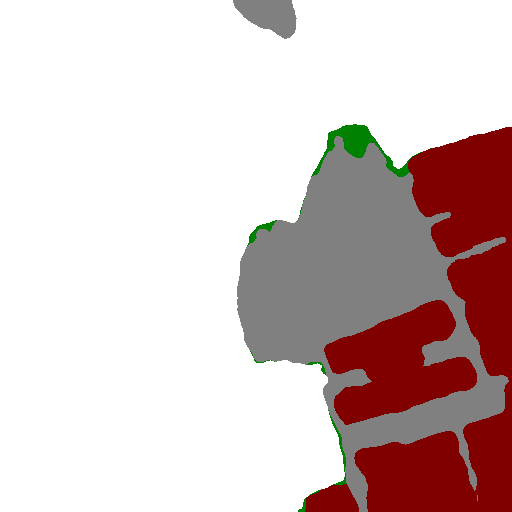} \\
        & Test image & GT & ChangMamba & BiSRNet & SAM-SCD & M-CD & SCanNet & GSTM-SCD & ChessMamba (Ours)\\
    \end{tabular}
    \caption{Qualitative comparison between our ChessMamba and SOTA methods for SCD (SECOND).} \label{Fig.compare_SCD} 
\end{figure*}

In Fig. \ref{Fig.compare_SCD} we qualitatively compare our ChessMamba with the recent SOTA SCD methods on the SECOND dataset. In the industrial scenario in (a), methods like BiSRNet and SAM-SCD struggle with fuzzy boundaries between trees and low-vegetation regions. By contrast, ChessMamba accurately distinguishes these classes, demonstrating superior detail preservation for fine-grained semantics. In building-centric scenarios (c-d), our approach sustains crisp structural outlines while competitors (e.g., M-CD and GSTM-SCD) exhibit blurred or broken boundaries. These representative examples reveal persistent challenges in existing methods, including semantic class bleeding and oversmoothed transitions. In contrast, our ChessMamba achieves geometrically precise object boundaries, reliable land-cover discrimination, and robust semantic change identification in complex environments.

\begin{figure*}[t]
\centering
    \includegraphics[width=0.8\linewidth]{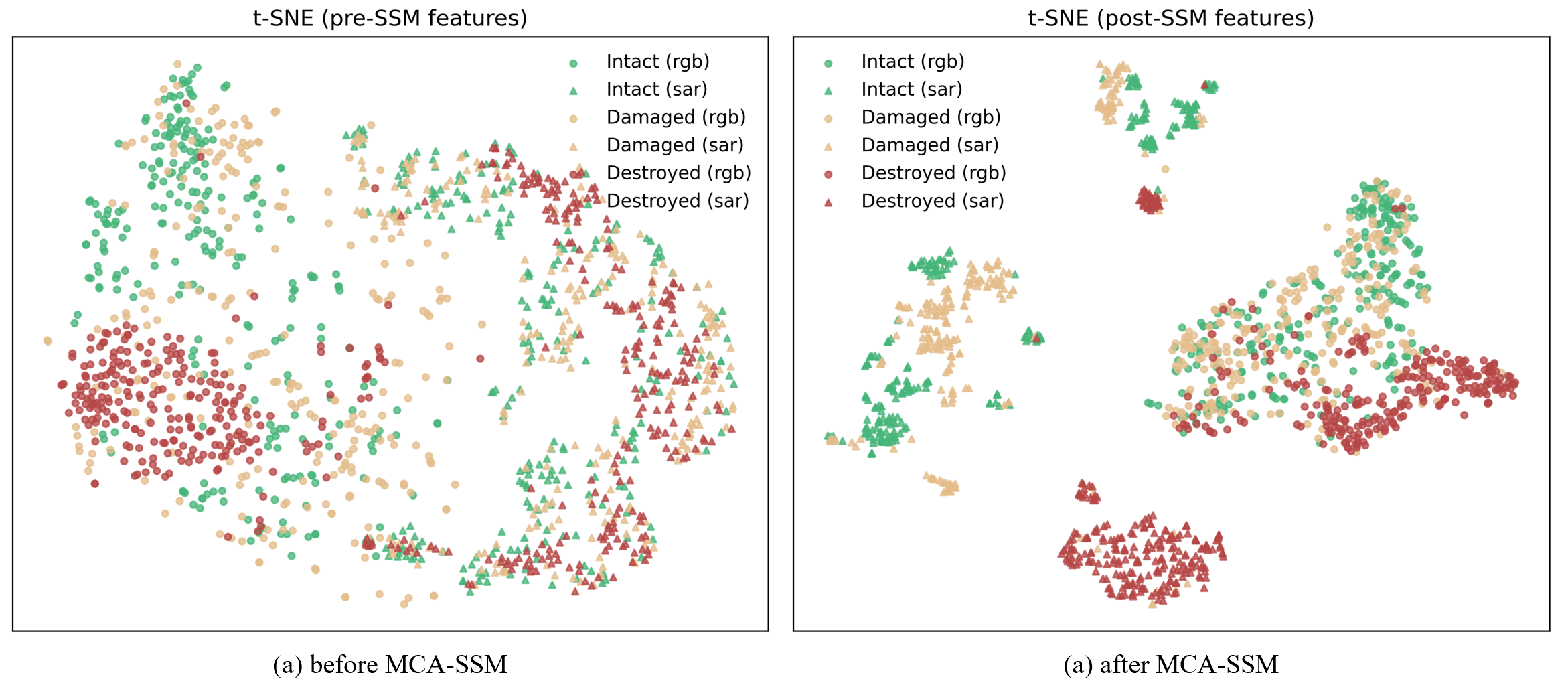}
    \caption{T-SNE visualization of the features before and after applying MCA-SSM (BRIGHT).}
\label{fig.vis_feat}
\end{figure*}

\begin{figure}[t]
\centering
    \includegraphics[width=1\linewidth]{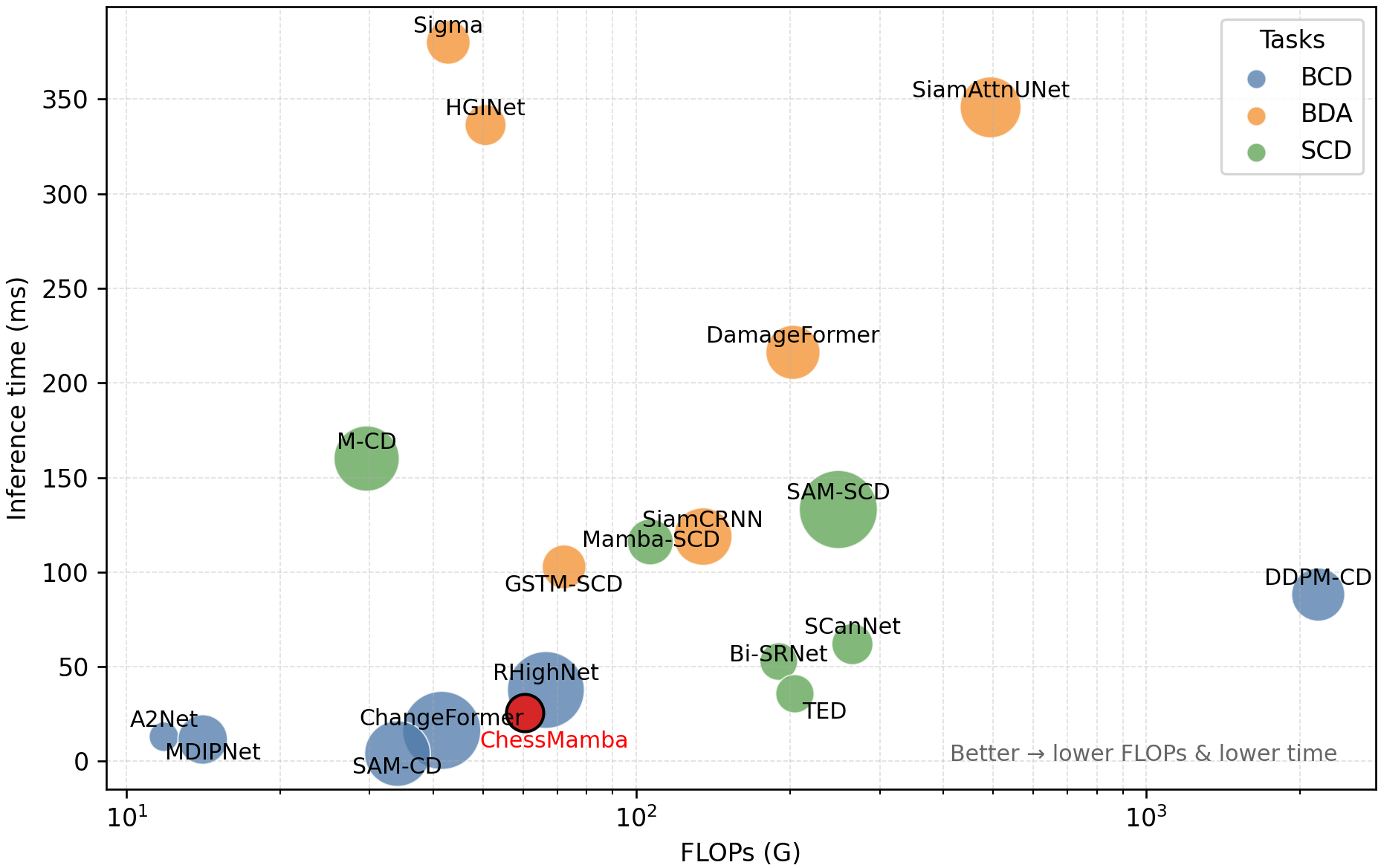}
    \caption{FLOPs vs. inference time of different CD methods. Marker size encodes parameter size, colors indicate task types. ChessMamba is highlighted in red.}
\label{fig.efficiency}
\end{figure}

\textbf{Efficiency Evaluation:}  Fig. \ref{fig.efficiency} summarizes three complementary aspects of computational cost for each method. The parameter amount (Mb), FLOPs (G) and inference time (ms) of each method are measured with the fixed spatial resolution of $512\times512$ pixels. Across BCD, BDA and SCD methods, we observe that several methods achieve strong accuracy at the price of heavy computation. For instance, transformers and diffusion-based approaches (e.g., HGINet, DDPM-CD) generally consume more GFLOPs and exhibit relatively high inference times, while some compact CNN baselines trade accuracy for speed. Our ChessMamba model occupies a distinct region of the efficiency spectrum: with approximately 22.67M parameters and 60.43G FLOPs, it attains one of the smallest computational footprints among competitive methods, and its inference time ($\sim$25 ms) is consistently lower than that of most recent baselines on all three tasks. This indicates that the proposed chessboard-structured Mamba decoder not only improves accuracy, but also delivers a favorable accuracy–efficiency trade-off that is well suited to high-resolution and time-critical CD scenarios.

Table \ref{table.efficiency} details computational efficiency statistics for diverse CD methods, expanding the analysis in Fig.\ref{fig.efficiency}. Specifically, the efficiency of ChessMamba is evaluated based on its variants corresponding to the different tasks (see Sec. \ref{sc3.tasks}).

In BCD, ChessMamba achieves competitive parameter efficiency, closely trailing A2Net (an efficient BCD model), while maintaining moderate FLOPs and inference time. For BDA, it surpasses all competitors in inference speed (23.02ms), despite moderately higher parameters (41.78M) than leading methods like HGINet,  and has efficient FLOPs (60.43G) relative to alternatives. In SCD, ChessMamba demonstrates superior parameter reduction (22.65M, minimal among all) and the fastest inference time, with FLOPs remaining competitive (second only to M-CD). Overall, ChessMamba demonstrates modest parameters, yet has notably accelerated inference across all tasks, and competitive computational demands.

\begin{table}[h]
\centering
\caption{Quantitative results of computational efficiency.}
\label{table.efficiency}
\resizebox{1\linewidth}{!}{%
\begin{tabular}{lccc}
\toprule
\textbf{Methods} & \textbf{Params (m)} & \textbf{FLOPs (G)} & \textbf{Inf. time (ms)} \\
\midrule
\multicolumn{4}{c}{\textbf{BCD Task}} \\
\midrule
ChangeFormer & 124.58 & 41.52 & 16.30 \\
A2Net & \textbf{14.42} & \textbf{11.83} & 12.94 \\
SAM-CD & 70.49 & 34.00 & \textbf{4.02} \\
MDIPNet & 40.21 & \underline{14.10} & 11.49 \\
STADE-CDNet & 47.31 & 48.77 & 12.46 \\
SEIFNet & 109.10 & 33.16 & \underline{7.86} \\
DDPM-CD & 46.41 & 2175.46 & 88.10 \\
RHighNet & 96.81 & 66.51 & 37.67 \\
ChessMamba (ours) & \underline{22.67} & 60.43 & 15.25 \\
\midrule
\multicolumn{4}{c}{\textbf{BDA Task}} \\
\midrule
SiamAttnUNet & 60.95 & 494.76 & 345.68 \\
SiamCRNN & 54.37 & 135.02 & \underline{118.78} \\
DamageFormer & 48.16 & 202.50 & 216.15 \\
HGINet & \textbf{27.70} & \underline{50.63} & 336.32 \\
Sigma & \underline{31.50} & \textbf{42.77} & 380.03 \\
ChessMamba (ours) & 41.78 & 60.43 & \textbf{23.02} \\
\midrule
\multicolumn{4}{c}{\textbf{SCD Task}} \\
\midrule
Bi-SRNet & \underline{23.39} & 189.91 & 52.70 \\
Mamba-SCD (tiny) & 34.69 & 106.64 & 115.94 \\
TED & 24.19 & 204.29 & \underline{35.65} \\
SCannNet & 27.90 & 264.95 & 61.92 \\
SAM-SCD & 117.16 & 248.61 & 133.07 \\
M-CD & 69.80 & \textbf{29.58} & 160 \\
GSTM-SCD & 31.55 & 72.21 & 102.76 \\
ChessMamba (ours) & \textbf{22.65} & \underline{60.47} & \textbf{22.74} \\
\bottomrule
\end{tabular}}
\end{table}

\section{Conclusion}

We present ChessMamba, a novel SSM framework that fundamentally rethinks cross-source feature fusion for CD in RS images. By integrating structural priors directly into selective state-space computations, our approach addresses critical limitations in existing SSM-based CD methods, particularly their vulnerability to spatiotemporal misalignment and fine-grained feature integration. The core innovations, including Chessboard interleaving with snake scanning and MCA-SSM for spatio-temporal contexts modeling, reshape multi-source interaction into a unified sequence while preserving 2D neighborhood topology.

Extensive experiments across BCD (Levir-CD), SCD (SECOND), and multimodal BDA (BRIGHT) demonstrate SOTA performance of our method. ChessMamba achieves substantial gains in localization precision and recall, notably reducing omission errors on Levir and enhancing damage classification under severe heterogeneity. This work establishes that embedding geometric consistency into state propagation is pivotal for efficient long-range modeling in change-sensitive applications. The proposed paradigms can potentially extends to tasks requiring structural coherence under misalignment.

\bibliographystyle{IEEEtran}
\bibliography{refs}

\end{document}